\documentclass[pmlr,twocolumn,10pt]{jmlr} 

\usepackage{booktabs}

\jmlrvolume{209}
\jmlrpages{350-378}
\jmlryear{2023}
\jmlrsubmitted{LEAVE UNSET}
\jmlrpublished{LEAVE UNSET}
\jmlrworkshop{Conference on Health, Inference, and Learning (CHIL) 2023} 

\newcommand{\repo}{\url{https://github.com/cavalab/proportional-multicalibration}}
\usepackage{wrapfig}
\usepackage{multirow}
\usepackage[noend]{algpseudocode}
\usepackage{algorithm,algorithmicx}

\usepackage[capitalize]{cleveref}
\Crefname{Definition}{Def.}{Def.}
\Crefname{definition}{Def.}{Def.}

\usepackage{clipboard}

\newcommand{\eps}{\varepsilon}
\renewcommand{\epsilon}{\varepsilon}
\newtheorem{claim}{Claim}

\newcommand{\A}{\mathcal{A}}

\newcommand{\C}{\mathcal{C}}
\newcommand{\G}{\mathcal{G}}

\newcommand{\X}{\mathcal{X}}
\newcommand{\card}[1] {\left\vert #1 \right\vert}
\newcommand{\Set}[1] {\left\{ #1 \right\}}

\newcommand{\EpstarI}{\E_D [ y | R \in I, x \in S]}
\newcommand{\Epstarr}{\E_D [ y | R = r, x \in S]}
\newcommand{\ERI}{\E_D [ R | R \in I, x \in S]}
\newcommand{\ERr}{\E_D [ R | R = r, x \in S]}
\DeclareMathOperator*{\E}{\mathbb{E}}

\title[Proportional Multicalibration]{
    Fair admission risk prediction with proportional multicalibration
}

\author{%
 \Name{William G. {La Cava}\nametag{\thanks{Corresponding author. \url{https://cavalab.org} \\
 Published version: \url{https://proceedings.mlr.press/v209/la-cava23a.html}}}} \Email{william.lacava@childrens.harvard.edu} \\
 \Name{Elle Lett} \Email{elle.lett@childrens.harvard.edu} \\
 \Name{Guangya Wan} \Email{gwan@hsph.harvard.edu}\\
 \addr Computational Health Informatics Program, Boston Children's Hospital, Harvard Medical School, Boston, MA, USA
}

\begin{document}

\maketitle
\begin{abstract}
Fair calibration is a widely desirable fairness criteria in risk prediction contexts. 
One way to measure and achieve fair calibration is with multicalibration. 
Multicalibration constrains calibration error among flexibly-defined subpopulations while maintaining overall calibration. 
However, multicalibrated models can exhibit a higher percent calibration error among groups with lower base rates than groups with higher base rates. 
As a result, it is possible for a decision-maker to learn to trust or distrust model predictions for specific groups. 
To alleviate this, we propose \emph{proportional multicalibration}, a criteria that constrains the percent calibration error among groups and within prediction bins. 
We prove that satisfying proportional multicalibration bounds a model's multicalibration as well its \emph{differential calibration}, a fairness criteria that directly measures how closely a model approximates sufficiency. 
Therefore, proportionally calibrated models limit the ability of decision makers to distinguish between model performance on different patient groups, which may make the models more trustworthy in practice.  
We provide an efficient algorithm for post-processing risk prediction models for proportional multicalibration and evaluate it empirically. 
We conduct simulation studies and investigate a real-world application of PMC-postprocessing to prediction of emergency department patient admissions.
We observe that proportional multicalibration is a promising criteria for controlling simultaneous measures of calibration fairness of a model over intersectional groups with virtually no cost in terms of classification performance.  
\end{abstract}

\paragraph*{Data and Code Availability}
This paper uses the MIMIC-IV-ED dataset
\citep{johnsonalistairMIMICIVED2021}, which is available on the PhysioNet repository
\citep{goldbergerPhysioBankPhysioToolkitPhysioNet2000}.
Code for the experiments is available here: \repo. 
\paragraph*{Institutional Review Board (IRB)}
This research does not require IRB approval. 


\section{Introduction}

Today, machine learning (ML) models have an impact on outcome disparities across sectors due to their wide-spread use in decision-making.  
When applied in clinical decision support (CDS), ML models help care providers decide whom to prioritize to receive finite and time-sensitive resources among a population of potentially very ill patients. 
These resources include hospital beds~\citep{barak-correnPredictionPatientDisposition2021,dinhOvercrowdingKillsHow2021}, organ transplants~\citep{schnellinger2021mitigating}, specialty treatment programs~\citep{henryTargetedRealtimeEarly2015,obermeyerDissectingRacialBias2019}, and, recently, ventilator and other breathing support tools to manage the COVID-19 pandemic~\citep{rivielloAssessmentCrisisStandards2022}. 

In scenarios like these, decision makers typically rely on risk prediction models to be \emph{calibrated}. 
Calibration measures the extent to which a model's risk scores, $R$, match the observed probability of the event, $P(y)$~\citep{brier1950verification}. 
Perfect calibration implies that $P(y|R=r) = r$, for all values of $r$. 
Calibration allows the risk scores to be used to rank patients in order of priority and informs care providers about the urgency of treatment. 
However, models that are not equally calibrated among subgroups defined by different sensitive attributes (race, ethnicity, gender, income, etc.) may lead to systematic denial of resources to marginalized groups (e.g.~\citep{obermeyerDissectingRacialBias2019,ashana2021equitably,roberts_fatal_2011,zelnick2021association,ku2021racial}). 
For example, ~\citet{obermeyerDissectingRacialBias2019} analyzed a large health system algorithm used to enroll high-risk patients into care management programs and showed that, at a given risk score, Black patients exhibited significantly poorer health than white patients. 

When evidence of algorithmic bias is observed, for example in estimates of kidney function~\citep{diao2021clinical}, it is unclear how care should be adjusted for affected patient groups. 
Variance in clinical judgements that are not evidence-based are themselves a source of unfairness, referred to as unwarranted clinical variation~\citep{harrisonAddressingUnwarrantedClinical2019,sutherlandUnwarrantedClinicalVariation2020,newhouseGeographicVariationMedicare2013}. 
Ideally, patients with the same evidence, e.g. model risk prediction, would receive the same standard of care. 
In this work, we propose measures and methods designed to reduce unwarranted variation in care that may arise from biased prediction models. 

To address equity in calibration, \citet{hebert-johnsonMulticalibrationCalibrationComputationallyIdentifiable2018} proposed a fairness measure called \textit{multicalibration} (MC), which asks that calibration be satisfied simultaneously over many flexibly-defined subgroups. 
Remarkably, MC can be satisfied efficiently by post-processing risk scores without negatively impacting the generalization error of a model, unlike other fairness concepts like demographic parity~\citep{fouldsAreParityBasedNotions2020} and equalized odds~\citep{hardtEqualityOpportunitySupervised2016a}. 
This has motivated the use of MC in practical settings (e.g.~\citet{bardaAddressingBiasPrediction2021a}) 
and has spurred several extensions~\citep{kimMultiaccuracyBlackboxPostprocessing2019,jungMomentMulticalibrationUncertainty2021,guptaOnlineMultivalidLearning2021,gopalanLowDegreeMulticalibration2022}.
If we bin our risk predictions, the MC criteria specifies that, for every group within each bin, the absolute difference between the mean observed outcome and the mean of the predictions should be small. 

As~\citet{barocasFairnessMachineLearning2019} note, fair calibration reduces to a more general fairness notion dubbed \emph{sufficiency}. 
Under sufficiency, the expected outcome should be independent of group membership, conditioned on the risk prediction. 
We see this criteria as highly desirable in CDS contexts.
Sufficiency eliminates a source of uncertainty from decision-making: how to interpret a model recommendation from a model that has been observed to perform more or less well on certain subpopulations.    
Given a risk score from a model satisfying sufficiency, a decision maker cannot distinguish between patient outcomes on the basis of their group membership. 
Thus, they cannot justify variations from the model's recommendations on the basis of patient identity alone.

In this work, we start by assessing the conditions under which MC satisfies sufficiency. 
To do so, we derive a fairness criteria directly from sufficiency, \emph{differential calibration} (DC).  
DC is an extension of differential fairness~\citep{foulds_intersectional_2019}, both named due to their relation to differential privacy~\citep{dworkAlgorithmicFoundationsDifferential2013a}.
DC constrains ratios of population risk between groups within risk prediction bins. 
We show that DC measures the extent to which one satisfies sufficiency in an interpretable way. 
In short, among patients assigned the same risk score from a model satisfying $\eps$-DC, the outcome is at most $e^{\eps}$ more likely in one group compared to any another. 
A low $\eps$-DC thereby constrains the amount by which a decision maker may learn to unequally trust the model for different groups.  

By relating sufficiency to MC, we describe a shortcoming of MC that can occur when the outcome probabilities are strongly tied to group membership. 
Under this condition, the amount of calibration error \emph{relative to the expected outcome} can be unequal between groups. 
This inequality hampers the ability of MC to (approximately) guarantee sufficiency except by setting extremely low error thresholds, resulting in the need for large numbers of updates.   

We propose a simple variant of MC called \textit{proportional multicalibration} (PMC) that instead requires the proportion of calibration error within each bin and group to be small. 
We prove that PMC bounds both multicalibration and differential calibration. 
We show that PMC can be satisfied with an efficient post-processing method, similarly to MC~\citep{pfistererMcboostMultiCalibrationBoosting2021}. 
Proportionally multicalibrated models thereby obtain robust fairness guarantees that are less dependent on population risk categories. 
It does so in fewer steps than MC by prioritizing updates to groups with high proportional calibration error.   

Finally, we investigate the application of these methods to predicting patient admissions in the emergency department, a real-world resource allocation task that is targeted by current CDS models \citep{barak-correnPredictionPatientDisposition2021}. 
We create a benchmark dataset for this task using the recently released MIMIC-IV emergency department dataset~\cite{johnsonalistairMIMICIVED2021}, and benchmark PMC- and MC-based postprocessing approaches.  
We show that post-processing for PMC results in models that are accurate, multicalibrated, and differentially calibrated.

\section{Reconciling Multicalibration and Sufficiency}
\label{s:methods}

\subsection{Preliminaries}

\looseness=-1
We consider the task of training a risk prediction model for a population of individuals with  outcomes, $y \in \Set{0,1}$, and features, $x \in \X$.  
Let $D$ be the joint distribution from which individual samples $(y, x)$ are drawn. 
We assume the outcomes $y$ are random samples from underlying independent Bernoulli distributions, denoted as $p^*(x) \in [0,1]$.
Individuals can be further grouped into \emph{collections of subsets}, $\C \subseteq \text{2}^{\X}$, such that $S \in \C$ is the subset of individuals belonging to $S$, and $x \in S$ indicates that individual $x$ belongs to group $S$. 
We denote our risk prediction model as $R(x): \X$ $\rightarrow [0,1]$.

In order to consider calibration in practice, the risk predictions are typically discretized into bins. 
We represent discretization by a parameter, $\lambda$, that specifies the width of the bins. 
As an example, $\lambda=0.1$ corresponds to decile bins. 
For brevity, proofs and some formal definitions in the following sections are given in~\appendixref{s:proof,s:app:def}.

\subsection{Multicalibration}
\looseness=-1
\citet{hebert-johnsonMulticalibrationCalibrationComputationallyIdentifiable2018} define multicalibration by first defining calibration with respect to a subset (i.e., group) of individuals: 

\begin{definition}[$\alpha$-calibration ]
\label{def:calibration}
Let $S \subseteq \X$.  
For $\alpha \in [0,1]$, $R$ is $\alpha$-\emph{calibrated} with respect to $S$ 
if there exists some $S' \subseteq S$ with $\card{S'} \ge (1-\alpha)\card{S}$
such that for all $r \in [0,1]$,
$$
\card{ \E_D [ y | R = r, x \in S' ] - r} \le \alpha.
$$
\end{definition}
MC then guarantees that $\alpha$-calibration holds over every subset from a collection of subsets: 

\begin{definition}[$\alpha$-Multicalibration]
\label{def:MC}
Let $\C \subseteq 2^{\X}$ be a collection of subsets of $\X$, $\alpha \in [0,1]$. 
A predictor $R$ is $\alpha$-multicalibrated on $\C$ if for all $S \in \C$,
$R$ is $\alpha$-calibrated with respect to $S$.
\end{definition}

We note that, according to \definitionref{def:calibration}, a model need only be calibrated over a sufficiently large subset of each group ($S'$) in order to satisfy the definition. 
This relaxation is used to maintain a satisfactory definition of MC when working with discretized predictions.
For simplicity, we conduct most of our analysis using the continuous versions of fairness definitions like \definitionref{def:MC} (see \appendixref{s:app:discrete} for an extended discussion). 

MC is one of few approaches to achieving fairness that does not require a significant trade-off to be made between a model's generalization error and the improvement in fairness it provides. 
As \cite{hebert-johnsonCalibrationComputationallyIdentifiableMasses2018} show, this is because achieving multicalibration is not at odds with achieving accuracy in expectation for the population as a whole. 
This separates calibration fairness from other fairness constraints like demographic parity and equalized odds~\citep{hardtEqualityOpportunitySupervised2016a}, both of which may denigrate the performance of the model on specific groups~\citep{chouldechovaFairPredictionDisparate2017a,pleissFairnessCalibration2017}. 
In clinical settings, such trade-offs may be difficult or impossible to justify. 
In addition to its alignment with accuracy in expectation, \citet{hebert-johnsonCalibrationComputationallyIdentifiableMasses2018} propose an efficient post-processing algorithm for MC similar on boosting. 
We discuss additional extensions to MC in \appendixref{s:app:related}. 

\subsection{Measuring Sufficiency via Differential Calibration}

MC provides a sense of fairness by approximating \emph{calibration by group}, which is perfectly satisfied when $P_D(y|R=r,x \in S)=r$ for all $S \in C$ and $r \in [0,1]$.
Calibration by group is closely related to the \emph{sufficiency} fairness criterion~\citep{barocasFairnessMachineLearning2019}. 
Sufficiency states that the outcome probability is independent from $\C$, conditioned on the risk score.  
In the binary group setting ($\C = \{S_i, S_j\}$), we can express sufficiency as
\begin{equation}\label{eq:sufficiency}
P_D(y|R, x \in S_i) = P_D(y|R, x \in S_j)
\end{equation}
or equivalently, 
$$
{ P_D(y|R, x \in S_i) }/{ P_D(y|R, x \in S_j) } = 1. 
$$

Unlike calibration by group, sufficiency does not stipulate that the risk scores be calibrated, yet from a fairness perspective, sufficiency and calibration by group are equivalent~\citep{barocasFairnessMachineLearning2019}.
In both cases, the sense of \emph{fairness} stems from the desire for $R$ to capture everything about group membership that is relevant to predicting $y$.

Under sufficiency, the risk score is equally informative of the outcome, regardless of group membership. 
Because of this, given a risk score from a model satisfying sufficiency, a decision maker cannot distinguish between patient outcomes on the basis of their group membership. 
In this sense, a model satisfying sufficiency provides an added level of trust in deployment: we know that a decision maker cannot justify different decisions on the basis of the patient's identity for the same risk prediction. 
If risk prediction models satisfied sufficiency, it would eliminate the need for group-specific decision protocol given recommendations from the same model. 
Consider, for example, the decision-making uncertainty related to estimates of kidney function~\citep{diao2021clinical}.  

Below, we define an approximate measure of sufficiency that constrains pairwise differentials between groups, and accomodates binned predictions: 

\begin{definition}[$\eps$-Differential Calibration]\label{def:DC}
    Let $\C \subseteq \text{2}^{\X}$ be a collection of subsets of $\X$. 
    A model $R(x)$ is $\eps$-differentially calibrated with respect to $\C$ if, 
    for all pairs $(S_i,S_j) \in \C \times \C$ for which $P_D(S_i), P_D(S_j) >0$, 
    for any $r \in [0, 1]$,

    \begin{equation}\label{eq:DC}
        e^{-\eps} \leq 
            \frac{\E_D [ y | R = r, x \in S_i]}
            {\E_D [ y | R = r, x \in S_j ]} 
        \leq e^{\eps}
        .
    \end{equation}
\end{definition}

By inspection we see that $\epsilon$ in $\epsilon$-DC measures the extent to which $R$ satisifies sufficiency. 
That is, when $P(y|R=r, x \in S_i) \approx P(y|R =r, x \in S_j)$ for all pairs $(S_i, S_j)$, $\eps \approx 0$. 
$\eps$-DC requires that, for any risk score, the outcome $y$ is at most $e^{\eps}$ times more likely among one group than another, and a minimum of $e^{-\eps}$ less likely. 

\definitionref{def:DC} fits the general definition of a \emph{differential fairness} measure proposed by~\citet{fouldsIntersectionalDefinitionFairness2019} and previously used to study demographic parity criteria~\citep{fouldsAreParityBasedNotions2020}. 
We describe the relation in more detail in~\appendixref{s:app:DF}, including \cref{eq:DC}'s connection to differential privacy~\citep{dwork2009differential} and pufferfish privacy~\citep{kiferPufferfishFrameworkMathematical2014}. 

Taken alone, DC does not prevent a decision-maker from equally distrusting a model for all patients.
That is because it does not guarantee the model is globally calibrated . 
Rather, it makes it harder to distinguish the calibration quality of the model between groups. 

\subsection{The differential calibration of multicalibrated models is limited by low-risk groups}

At a basic level, the form of MC and sufficiency differ: MC constrains absolute differences between groups across prediction bins, whereas sufficiency constrains pairwise differentials between groups. 
To reconcile MC and DC/sufficiency more formally, we pose the following question: if a model satisfies $\alpha$-MC, what, if anything does this imply about the $\eps$-DC of the model?   
(In \appendixref{s:app:thm},\theoremref{thm:DCtoMC}, we answer the inverse question).
We now show that multicalibrated models have a bounded DC, but that this bound is limited by small values of $R$. 
\begin{theorem}
\label{thm:MCtoDC}
\Copy{thm:MCtoDC}{
    Let $R(x)$ be a model satisfying $\alpha$-MC on a collection of subsets $\C \in 2^{\X}$. 
    Let $r_{min} = \min_{S \in \C}{\ERr}$ be the minimum expected risk prediction among $S \in \C$ and $r \in [0,1]$. 
    Then R(x) is $\left( \ln \frac{r_{min}+\alpha}{r_{min}-\alpha} \right)$-differentially calibrated. 
}
\end{theorem}

\theoremref{thm:MCtoDC} illustrates the important point that, \emph{in terms of percentage error}, $MC$ does not provide equal protection to groups with different risk profiles. 
Imagine a model satisfying (0.05)-MC for groups $S \in \C$. 
Consider individuals receiving model predictions $R(x) = 0.9$. 
MC guarantees that, for any category $\Set{x : x \in S, R(x) = 0.9}$, the expected outcome probability is at least $0.9 - \alpha = 0.85$ and at most $0.9 + \alpha = 0.95$. 
This bounds the percent error among groups with this prediction to about 6\%.
In contrast, consider individuals for whom $R(x) = 0.3$; each group may have a true outcome probability as low as 0.25, which is an error of 20\% - about 3.4x higher than the percent error in the higher-risk group. 

In Appendix \theoremref{thm:DCtoMC}, we show that differentially calibrated models can bound multicalibration only if they are also $\alpha$-calibrated (\definitionref{def:calibration}). 

\section{Proportional Multicalibration} 

\begin{figure}
    \includegraphics[width=\columnwidth]{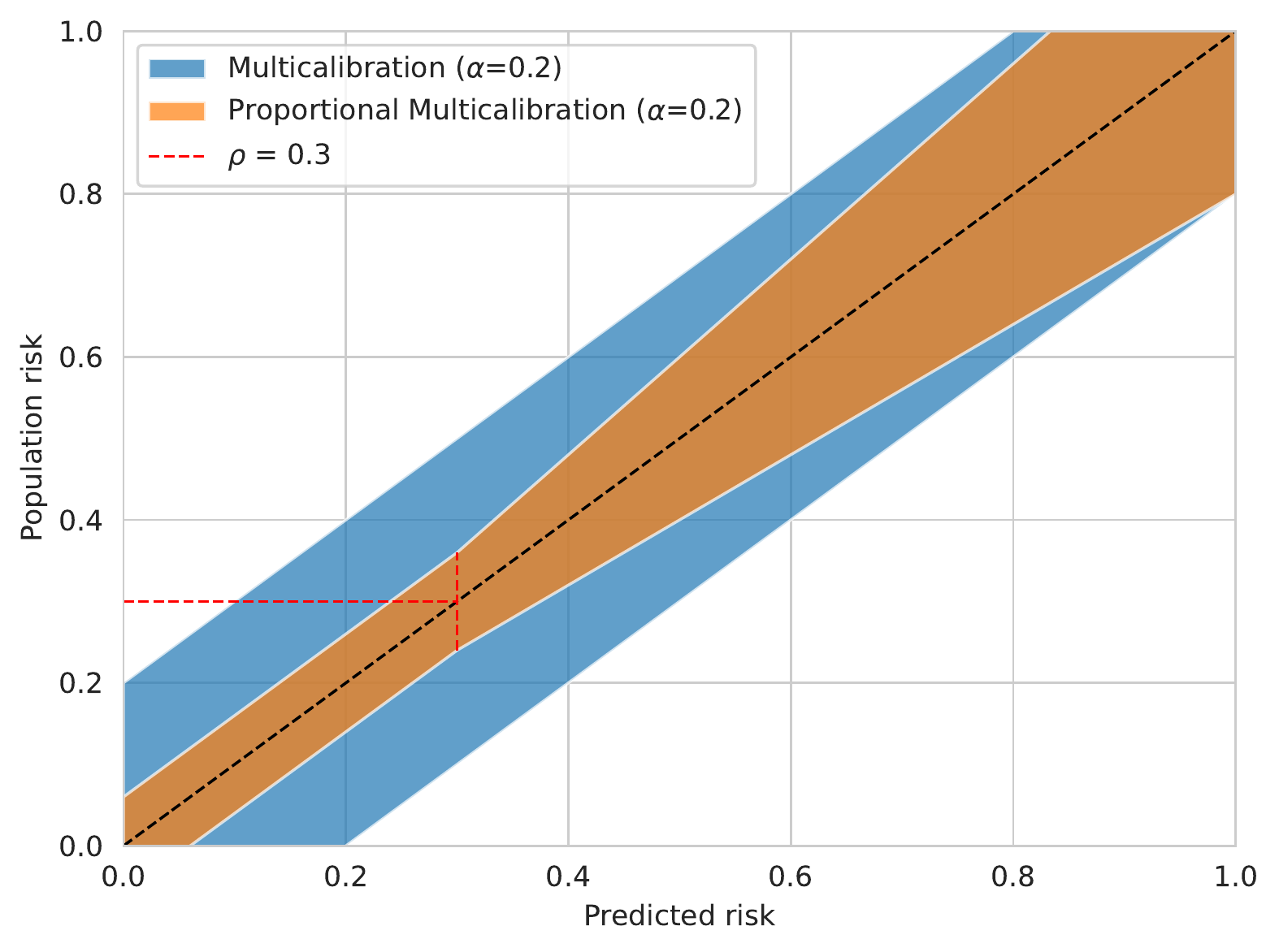}
    \caption{Visualization of the constraint differences between multicalibration and proportional multicalibration (PMC). 
    The filled area represents the maximum the predicted risk can deviate from the fraction of positives in the population, for any group. 
    Below $\rho$, the constraint is constant for PMC. 
    }
    \label{fig:constraints}
\end{figure}

We are motivated to define a measure that is efficiently learnable like MC (\definitionref{def:MC}) but better aligned with the multiplicative interpretation of sufficiency, like DC (\definitionref{def:DC}). 
To do so, we define PMC, a variant of MC that constrains the proportional calibration error of a model among subgroups and risk strata. 
In this section, we show that bounding a model's PMC is enough to meaningfully bound DC and MC. 
Furthermore, we provide an efficient algorithm for satisfying PMC based on a simple extension of MC/Multiaccuracy boosting~\citep{kimMultiaccuracyBlackboxPostprocessing2019}.
We begin by defining proportional calibration, which expresses calibration error as a percentage of the outcome probability among a group. 

\begin{definition}[$\alpha$-Proportional Calibration]\label{def:PC}
\label{def:prop_calibration}
Let $S \subseteq \X$.  
For $\alpha > 0$, $R$ is $\alpha$-\emph{proportionally calibrated} with respect to $S$ 
if there exists some $S' \subseteq S$ with $\card{S'} \ge (1-\alpha)\card{S}$
such that for all $r \in [0,1]$,
$$
\card{ \E_D [ y | R = r, x \in S' ] - r} \le \alpha \; \E_D [ y | R = r, x \in S' ].
$$
\end{definition}

Proportional multicalibration is then defined by requiring \definitionref{def:prop_calibration} be satisified among a collection of groups: 

\begin{definition}[$\alpha$-Proportional Multicalibration]\label{def:PMC}
Let $\C \subseteq 2^{\X}$ be a collection of subsets of $\X$, $\alpha > 0$. 
A predictor $R$ is $\alpha$-\emph{proportionally multicalibrated} on $\C$ if for all $S \in \C$,
$R$ is $\alpha$-proportionally calibrated with respect to $S$.
\end{definition}

We also define a discretized version of PMC in \appendixref{s:app:def} that is useful for implementing the measure in \algorithmref{alg:PMC} and measuring PMC in our experiments. 
In \appendixref{s:app:discrete}, we show  $(\alpha,\lambda)$-PMC meaningfully bounds $\alpha$-PMC under different discretizations, such that we can minimize $(\alpha,\lambda)$-PMC to achieve low $\alpha$-PMC. 

In practice, we must ensure that the outcome probability for any (group, prediction bin) category is greater than zero for PMC to be meaningful. 
We later introduce a lower bound, $\rho$, to prevent the outcome probability from being too small.  
It is common in clinical settings to only show a risk prediction to a decision maker if it exceeds some threshold; in those settings, $\rho$ can be set to match this threshold.  

\paragraph{Comparison to Differential Calibration}
Rather than constraining the differentials of prediction- and group- specific outcomes among all pairs of subgroups in $\C \times \C$ as in DC (\definitionref{def:DC}), PMC constrains the relative error of each group in $\C$. 
In practical terms, this makes it more efficient to calculate PMC by a factor of $O(\card{\C})$ steps compared to DC.  
In addition, PMC constrains each group's calibration with respect to ground truth, whereas DC only constrains the differentials \textit{between} groups. 
We formalize the relationship between these two measures below. 

\begin{theorem}
\label{thm:PMCtoDC}
\Copy{thm:PMCtoDC}
{
    Let R(x) be a model satisfying $(\alpha)$-PMC on a collection $\C$. 
    Then $R(x)$ is $(\ln \frac{1+\alpha}{1-\alpha})$-differentially calibrated.
}
\end{theorem}

\theoremref{thm:PMCtoDC} demonstrates that $\alpha$-proportionally multicalibrated models satisfy a straightforward notion of differential fairness that depends monotonically only on $\alpha$. 
Because PMC bounds DC, proportionally calibrated models inherit desirable properties of differentially calibrated models, especially the following: 

\begin{corollary}
    \label{corr:private}
    Let R(x) be a model satisfying $(\alpha)$-PMC on a collection $\C$. 
    Then $R(x)$ satisfies $(\ln \frac{1+\alpha}{1-\alpha})$-pufferfish privacy with respect to $\C$.
\end{corollary}

\corollaryref{corr:private} follows from the definitions of $\eps$-DC and pufferfish privacy (Appendix \definitionref{def:puff},~\cite{kiferPufferfishFrameworkMathematical2014}).
This property is perhaps best understood as a guarantee of trust rather than privacy. 
Namely, it guarantees that the patient's group identity does not provide useful information to a decision-maker in deciding how much to trust model predictions. 
The patient’s group identity is in this sense “private” from the model’s calibration. 
Unlike many privacy-preserving algorithms, we do not need to add add any noise to the model to satisfy DC; conversely, it is achieved by making the model better calibrated for some groups.

\paragraph{Comparison to Multicalibration}
Rather than constraining the absolute difference between risk predictions and the outcome as in MC, PMC requires that the calibration error be a small fraction of the expected risk in each category $(S,I)$.  
In this sense, it provides a stronger protection than MC by requiring calibration error to be a small fraction regardless of the risk group.  
In many contexts, we would argue that this is also more aligned with the notion of fairness in risk prediction contexts.  
Under MC, the underlying probability of an outcome within a group affects the fairness protection that is received (i.e., the percentage error that \definitionref{def:MC} allows).  
Because underlying probabilities of many clinically relevant outcomes vary significantly among subpopulations, multicalibrated models may systematically permit higher percentage error to specific groups. 
The difference in relative calibration error among populations with different risk profiles also translates in weaker sufficiency guarantees, as demonstrated in~\theoremref{thm:MCtoDC}.  
In contrast, PMC provides a fairness guarantee that is less dependent on subpopulation risks.  

In the following theorem, we show that MC is also constrained when a model satisfies PMC. 

\begin{theorem}
    \label{thm:PMCtoMC}
    \Copy{thm:PMCtoMC}{
    Let $R(x)$ be a model satisfying \emph{$\alpha$-PMC} on a collection $\C$. 
    Then $R(x)$ is ($\frac{\alpha}{1-\alpha}$)-multicalibrated on $\C$. 
    }
\end{theorem}

The proof of \theoremref{thm:PMCtoMC} is given in \appendixref{proof:PMCtoMC}. 
This theorem implies that a proportionally calibrated model with sufficiently low $\alpha$ will satisfy a similarly low value of MC.
We further discuss and illustrate the bounds given by \theoremref{thm:MCtoDC,thm:PMCtoDC,thm:PMCtoMC} in \appendixref{s:app:illustrate}.  

Because PMC bounds a model's multicalibration, proportionally multicalibrated models also bound the generalization error of a predictor such that it is close to the ``best in class'' predictions. 

\begin{corollary}
    \label{corr:generalization}
    Let R(x) be a model satisfying $(\alpha)$-PMC on a collection $\C$. 
    Let $\mathcal{H}$ be a set of predictors, $p^*$ the outcome-generating distribution, and $h^* = \arg\min_{h \in \mathcal{H}} || h - p^* ||^2$. 
    Then  
    $$
    || R - p^* ||^2 - || h^* - p^* ||^2 < \frac{6 \alpha}{1-\alpha}. 
    $$
\end{corollary}

This guarantee follows directly from Theorem 5 in \cite{hebert-johnsonMulticalibrationCalibrationComputationallyIdentifiable2018}. 
In a nutshell, it means that proportionally multicalibrated models are bound to be close to the best predictor of their hypothesis class ($\mathcal{H}$).  
This contrasts fair calibration with other approaches to fairness that typically form a trade-off with a model's overall error. 
We further illustrate relationship between PMC, DC, and MC in Appendix \cref{fig:params}. 

\subsection{Learning proportionally multicalibrated predictors}
\label{s:pmcboost}

In \algorithmref{alg:PMC}, we propose an extension of MCBoost~\citep{pfistererMcboostMultiCalibrationBoosting2021} to efficiently update risk predictors to satisfy PMC. 
\algorithmref{alg:PMC} works by checking for calibration errors among groups and prediction intervals that violate the user threshold, and adjusting these predictions towards the target. 
PMCBoost differs in two main ways: first, it updates whenever calibration error is not within $\alpha \bar{y}$ for all categories, as opposed to simply within $\alpha$. 
Second, it ignores updates for categories with low outcome probability (less than $\rho$).  
Next, we prove that PMCBoost learns an ($\alpha$,$\lambda$)-PMC model in a polynomial number of steps.

\begin{proposition}
\label{prop:alg}
\Copy{prop:alg}{
    Define $\alpha, \lambda, \gamma, \rho > 0$. 
    Let $\C \subseteq 2^{\X}$ be a collection of subsets of $\X$ such that, for all $S \in \C$, $P_D(S) > \gamma$. 
    Let $R(x)$ be a risk prediction model to be post-processed. 
    For all 
    $(S,I) \in \C \times \Lambda_{\lambda}$, let 
    $E[y|R\in I, x \in S] > \rho$. 
    There exists an algorithm that satisfies $(\alpha, \lambda)$-PMC with respect to $\C$ 
    in $O(\frac{|C|}{\alpha^3\lambda^2\rho^2\gamma})$ steps.
}
\end{proposition}

We analyze \algorithmref{alg:PMC} and show it satisfies Proposition \ref{prop:alg} in \appendixref{s:proof:alg}.
This more stringent threshold requires an additional $O(\frac{1}{\rho^2})$ steps compared to the algorithm for MC, where $\rho>0$ is a lower bound on the expected outcome within a category $(S,I)$.  

\paragraph{ Achieving proportional multicalibration with MCBoost}

Another way to achieve $\alpha_P$-PMC is to use the existing MCBoost algorithm, but setting $\alpha_M = \rho \alpha_P$. 
In other words, setting a very low value for $\alpha_M$ should also satisfy \cref{def:PMC} because, 
if $\alpha_M$ can be made small enough, the calibration error 
on all categories will be small compared to the outcome prevalence, $\EpstarI$. 
However, to achieve PMC guarantees by MCBoost requires a large number of unnecessary updates for high risk groups, since the DC and PMC of multicalibrated models are limited by low-risk groups (\theoremref{thm:MCtoDC}). 
Furthermore, the number of steps in MCBoost (and PMCBoost scales as an inverse high-order polynomial of $\alpha$ (cf. Thm. 2~\citep{hebert-johnsonMulticalibrationCalibrationComputationallyIdentifiable2018}). 

We can compare the algorithm complexity of MCBoost and PMCBoost to get a better comparison. 
Say we would like to satisfy $\alpha$-PMC using MCBoost. 
\citet{hebert-johnsonMulticalibrationCalibrationComputationallyIdentifiable2018} show that MCBoost takes $O(\frac{|C|}{\alpha^3\lambda^2\gamma})$ steps to complete. 
We must reduce $\alpha$-MC by a factor of $\rho$ to satisfy $\alpha$-PMC; this increases the number of MCBoost steps by a factor of $(\frac{1}{\rho^3})$. 
For a reasonable value of $\rho$ (say, $\rho=0.1$), this takes 1000x more steps, and corresponds to 10x more steps than it would take to satisfy $\alpha$-PMC with PMCBoost.  

Now consider the reverse: satisfying $\alpha$-MC using PMCBoost. 
Because of \theoremref{thm:PMCtoMC}, we need only reduce $\alpha$-PMC by a factor of $\frac{1}{1+\alpha\text{-MC}}$, which increases the number of PMCBoost steps by $(1+\alpha\text{-MC})^3$.
For example, to achieve $\alpha$-MC = 0.1, PMCBoost takes about 1.3x more steps than MCBoost. 

\begin{algorithm2e}
  \DontPrintSemicolon
  \label{alg:PMC}
  \caption{PMCBoost}
  \scriptsize
  \KwIn{ 
    predictor $R(x)$; \\
    parameters $\alpha, \lambda, \gamma, \rho > 0$; \\
    $\C \in 2^{\X}$ such that for all $S \in \C, P_D(S) \geq \gamma$; \\
    $\mathcal{D} = \{(y,x)_i\}_{i=0}^{N} \sim D$
  }
  \KwOut{ Updated predictor $R(x)$}
  
  \Repeat{No updates to $R(x)$} 
  {
    $\{(y,x)\}$ sample $\mathcal{D}$ \;
    \For{$S \in \C, I \in \Lambda_\lambda$ 
        such that $P_D(R \in I , x \in S) \geq \alpha \lambda \gamma$ }
    {
      $S_r$ $\leftarrow$ $S \cap \Set{x: R(x) \in I}$ \;
      
      $\bar{r}$ $\leftarrow$ $\frac{1}{|S_r|}\sum_{x \in S_r}{R(x)}$ \Comment{average group prediction } \;

      $\bar{y}$ $\leftarrow$ $\frac{1}{|S_r|}\sum_{x \in S_r}{y(x)}$ \Comment{average subgroup risk} \;

      $\Delta r$ $\leftarrow$ $\bar{y} - \bar{r}$ \;

      \uIf{$\bar{y} \geq \rho$}
      {
        cutoff $\leftarrow$ $\alpha \bar{y}$
      }
      \Else(\Comment{limit low probability updates}){
        cutoff $\leftarrow$ $\alpha \rho$ 
      }

      \If{$\card{\Delta r} \geq$ cutoff }
      {
        $R(x)$ $\leftarrow$ $R(x) + \Delta r$ for all $x \in S_r$ \;
        $R(x)$ $\leftarrow$ squash($R(x)$, $[0,1]$) \Comment{limit to $[0,1]$}  \;
      }
    }
  }
\end{algorithm2e}

\section{Experiments}
\label{s:exp}

\begin{table*}
    \centering
    \scriptsize
    \caption{
        Admission prevalence (Admissions/Total (\%)) among patients in the MIMIC-IV-ED data repository, stratified by the intersection of ethnoracial group and gender. 
    }
    \label{tbl:mimic}
    \begin{tabular}{lrrr}
\toprule
Gender &                  F &                  M &             Overall \\
Ethnoracial Group             &                    &                    &                     \\
\midrule
American Indian/Alaska Native &       70/257 (27\%) &       82/170 (48\%) &       152/427 (36\%) \\
Asian                         &    1043/3595 (29\%) &    1032/2384 (43\%) &     2075/5979 (35\%) \\
Black/African American        &   3124/27486 (11\%) &   2603/14458 (18\%) &    5727/41944 (14\%) \\
Hispanic/Latino               &   1063/10262 (10\%) &    1168/5795 (20\%) &    2231/16057 (14\%) \\
Other                         &    1232/5163 (24\%) &    1479/3849 (38\%) &     2711/9012 (30\%) \\
Unknown/Unable to Obtain      &    1521/2156 (71\%) &    2074/2377 (87\%) &     3595/4533 (79\%) \\
White                         &  18147/50174 (36\%) &  18951/45435 (42\%) &   37098/95609 (39\%) \\
Overall                       &  26200/99093 (26\%) &  27389/74468 (37\%) &  53589/173561 (31\%) \\
\bottomrule
\end{tabular}

\end{table*}

\begin{table}
    \centering
    \scriptsize
    \caption{Features used in the hospital admission task.}
    \label{tbl:features}
    \begin{tabular}{p{8em} p{18em}}
        \toprule
        Description     &   Features   \\
        \midrule
        Vitals  &
            temperature, heartrate, 	resprate, 	o2sat, 	systolic blood pressure, 	diastolic blood pressure 	
            \\
        Triage Acuity &  
            Emergency Severity Index~\citep{tanabeReliabilityValidityScores2004}
        \\
        Check-in Data   &
            chief complaint, self-reported pain score
        \\
        Health Record Data  &
            no. previous visits, no. previous admissions 
        \\
        Demographic Data    &
            ethnoracial group, gender, age, marital status, insurance, primary language
        \\
        \bottomrule
    \end{tabular}
\end{table}

In our first set of experiments (\cref{s:exp}), we study MC and PMC in simulated population data to understand and validate the analysis in previous sections. 
In the second section, we compare the performance of varied model treatments on a real world hospital admission task, using an implementation of \algorithmref{alg:PMC}. 
We make use of empirical versions of our fairness definitions which we refer to as \textit{MC loss}, \textit{PMC loss}, and \textit{DC loss}.  
In short, these measures calculate the maximum (proportional) calibration error or pairwise calibration differential among subgroups and risk categories in the data sample.    
Due to space constraints the formal definitions are given in \appendixref{s:app:def} (\definitionref{def:mcloss,def:pmcloss,def:dcloss}).

\paragraph{Simulation study} 
\label{s:exp:sim}
We simulate data from $\alpha$-multicalibrated models. 
For simplicity, we specify a data structure with a one-to-one correspondence between subset  and model estimated risk, such that for all $x$ in $S$, $R(x)=R(x|x\in S)=R(S)$.
Therefore all information for predicting the outcome based on the features in $x$ is contained in the attributes $\mathcal{A}$ that define subgroup $S$. 
Outcome probability is specified as
$p_i^*=P_D(y|x \in S_i)=0.2+0.01(i-1)$
and $i=1,\cdots,N_s$, where $N_s$ is the number of subsets $S$, defined by $\mathcal{A}$ and indexed by $i$  with increasing $p^*$. 
For each group,
$R_i=R(S_i)=R(x|x \in S_i)=p_i^*-\Delta_i.$
We randomly select $\Delta_i$ for one group to be $\pm\alpha$ and for the remaining groups, $\Delta_i= \pm\delta$, where $\delta\sim \textrm{Uniform}(\min=0, \max=\alpha)$. 
In all cases, the sign of $\Delta_i$ is determined by a random draw from a Bernoulli distribution.  For these simulations we set $N_S=61$ and $\alpha=0.1$, such that $p^*_i\in[0.2,0.8]$ and $R_i\in[0.1,0.9]$. We generate $N_{sim}=1000$ simulated datasets, with $n=1000$ observations per group, and for each $S_i$, we calculate the ratio of the absolute mean error to $p^*_i$, i.e. the PMC loss function for this data generating mechanism. 

We also simulate three specific scenarios where: 
    1) $\card{\Delta_i}$ is equivalent for all groups (Fixed); 
    2) $\card{\Delta_i}$ increases with increasing $p_i^*$; and 
    3) $\card{\Delta_i}$ decreases with increasing $p_i^*$,
with $\alpha=0.1$ in each case.
These scenarios compare when $\alpha$ is determined by all groups, the group with the lowest outcome probability, and the group with the highest outcome probability, respectively.

\paragraph{Hospital admission}
\label{s:exp:mimic}
Next, we test PMC alongside other methods in application to prediction of inpatient hospital admission for patients visiting the emergency department (ED). 
Overcrowding and long wait times in EDs have been shown to increase odds of inpatient death (5\%, CI 2-8\%), length of stay (0.8\%, CI 0.5-1\%), and costs per admission (1\%, CI 0.7-2\%)~\citep{sunEffectEmergencyDepartment2013}. 
The burden of overcrowding and long wait times in EDs is significantly higher among non-white, non-Hispanic patients and socio-economically marginalized patients~\citep{jamesAssociationRaceEthnicity2005,mcdonaldExaminingAssociationCommunityLevel2020a}. 

Recent work has demonstrated risk prediction models that can expedite patient visits by predicting patient admission at an early stage of a visit with a high degree of certainty (AUC $\geq$ 0.9 across three large care centers)~\citep{barak-correnProgressivePredictionHospitalisation2017,barak-correnEarlyPredictionModel2017,barak-correnPredictionHealthcareSettings2021,barak-correnPredictionPatientDisposition2021}. 
Our goal is to ensure no group of patients will be over- or under-prioritized over another by these models, which could exacerbate the treatment and outcome disparities that currently exist.  

\begin{table}[t]
        \centering
        \scriptsize
        \caption{Parameters for the hospital admission prediction experiment.}
        \label{tbl:params}
        \begin{tabular}{p{12.5em} p{10em}}
\toprule
    Parameter   &   Values  \\
    \midrule
    tolerance   ($\alpha$)       &   (0.001, 0.01, 0.1, 0.2) \\
    min group probability  ($\gamma$) &   (0.05, 0.1) \\
    binning ($\lambda$)     &   0.1 \\
    min outcome probability ($\rho$) &   (0.01, 0.1) \\
    Base Model &   LR, RF, DNN \\
    Groups        & 
    (race/ethnicity, gender), (race/ethnicity, gender, insurance product)
    \\

    \bottomrule
\end{tabular}
\end{table}

We construct a prediction task similar to previous studies but using a new data resource: the \href{https://physionet.org/content/mimic-iv-ed/1.0/}{MIMIC-IV-ED} repository~\citep{johnsonalistairMIMICIVED2021}.
After data preparation (see \appendixref{s:app:exp}), the cohort consists of 173,561 visits with demographics and admission statistics given in~\cref{tbl:mimic}. 
\cref{tbl:features} shows the list of features used for prediction, modelled off prior work~\citep{barak-correnPredictionPatientDisposition2021}.
In \cref{tbl:mimic} we observe stark differences in admission rates by demographic group and gender, suggesting that the use of a proportional measure of calibration could be appropriate for this task. 
We trained and evaluated $\ell_1$-penalized logistic regression (LR), random forest (RF), and deep neural network (DNN) models of patient admission, with and without post-processing with MCBoost~\citep{pfistererMcboostMultiCalibrationBoosting2021} or PMCBoost. 
We varied $\alpha$, $\gamma$, and $\rho$ to characterize parameter sensitivity among the methods (\cref{tbl:params}). 
For each of the parameter settings, we conducted 100 repeat experiments with different shuffles of the data. 
Comparisons are reported on a test set of 20\% of the data for each trial.  
Additional experiment details are available in~\appendixref{s:app:exp}.

\section{Results}

\begin{figure}
        \includegraphics[width=\linewidth]{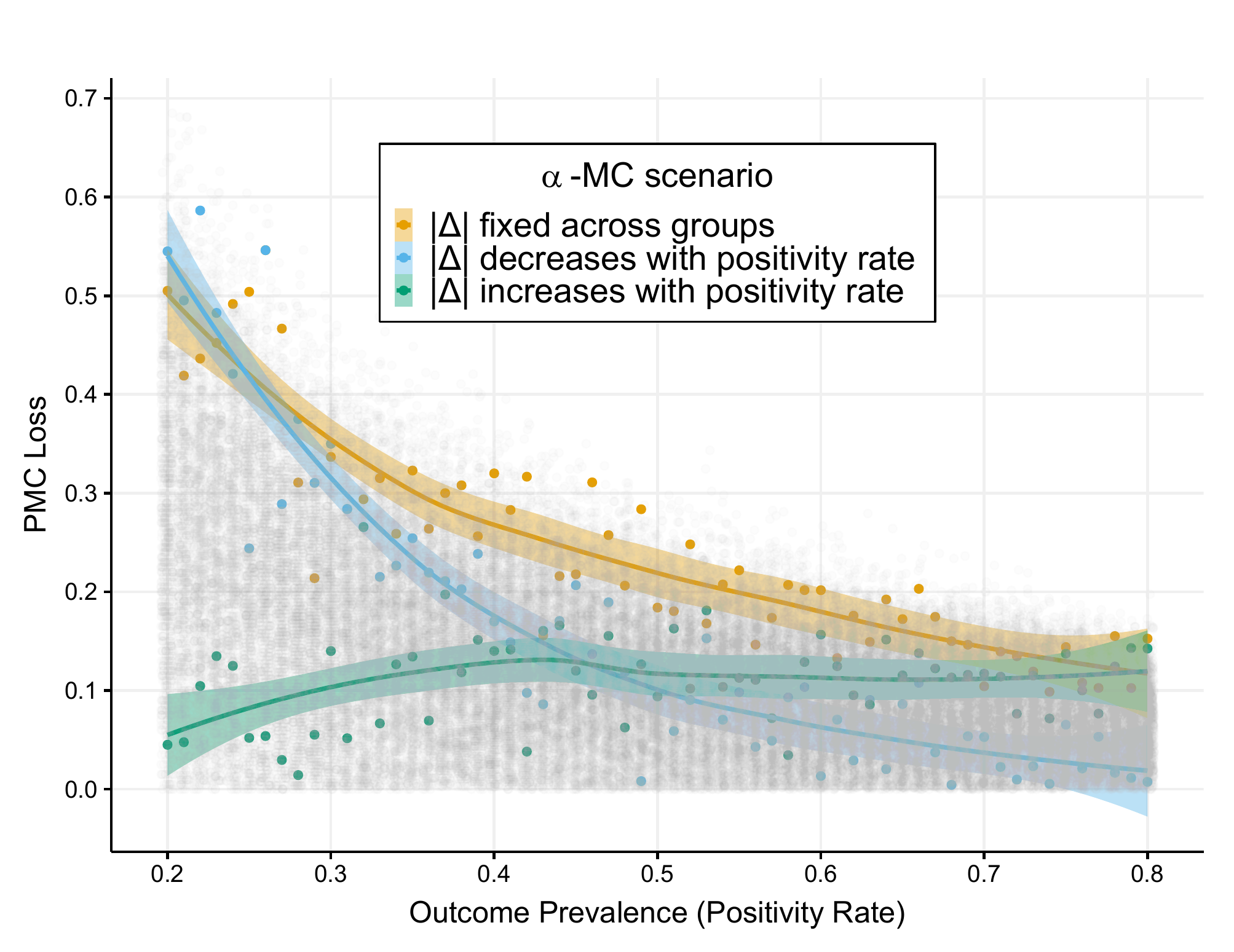}
        \caption{
            The relationship between MC, PMC, and outcome prevalence as illustrated via a simulation study in which the rates of the outcome are associated with group membership. 
            Gray points denote the PMC loss of a (0.1)-MC model on 1000 simulated datasets, and colored lines denote three scenarios in which each group's calibration error ($|\Delta|$) varies.
            Although MC is identical in all scenarios, PMC loss is higher among groups with lower positivity rates in most scenarios unless the groupwise calibration error increases with positivity rate. 
        }
    \label{fig:sim}
\end{figure}

\cref{fig:sim} shows the results of our simulation study. 
The results indicate that, without the proportionality factor, $\alpha$-multicalibrated models exhibit a dependence between the group prevalence and the amount of proportional calibration loss.
The results demonstrate why $\alpha$-MC alone is not sufficient to achieve sufficiency, particularly when outcome probabilities vary by group.  

\begin{figure*}
    \includegraphics[width=\linewidth]{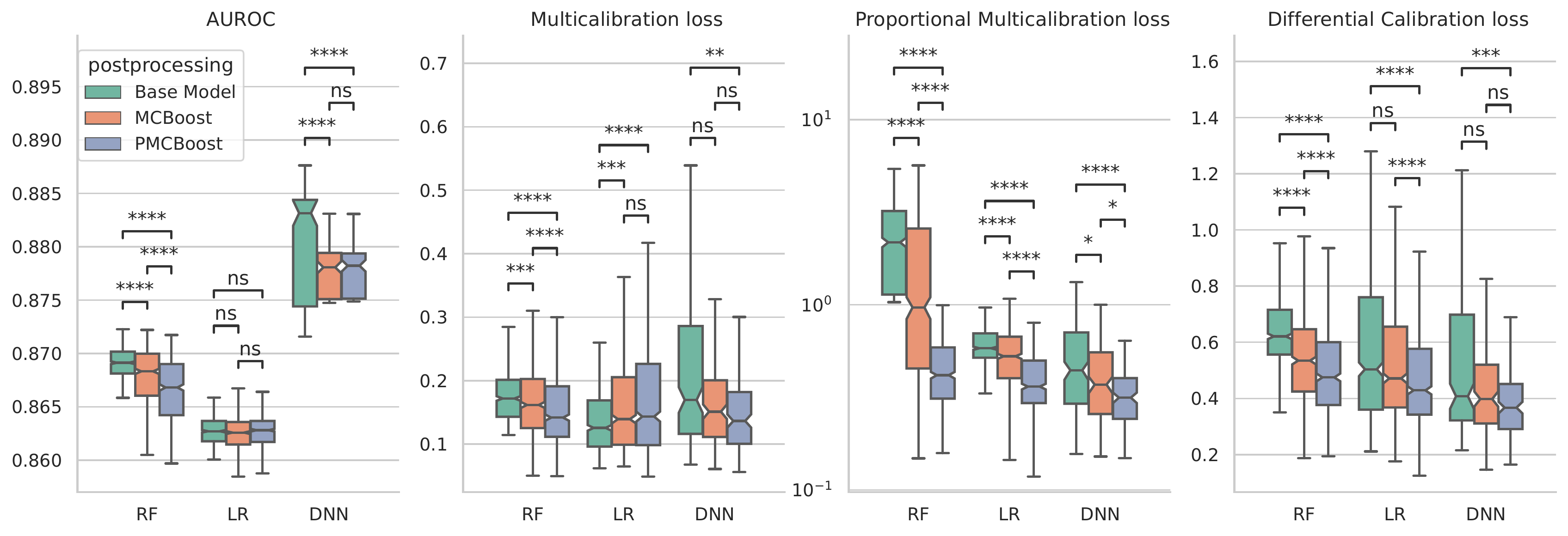}
    \caption{
        A comparison of LR, RF, and DNN models, with and without MCBoost and PMCBoost, on the hospital admission task. 
        From left to right, trained models are compared in terms of test set AUROC, MC loss, PMC loss, and DC loss. 
        Points represent the median performance over 100 shuffled train/test splits with bootstrapped 99\% confidence intervals. 
        We test for significant differences between post-processing methods using two-sided Wilcoxon rank-sum tests with Bonferroni correction. 
        ns: $p <=$ 1; 
      **: 1e-03 $< p <=$ 1e-02;
     ***: 1e-04 $< p <=$ 1e-03;
    ****: $p <=$ 1e-04.
    }
    \label{fig:mimic_results}
\end{figure*}

\begin{figure}
    \includegraphics[width=\columnwidth]{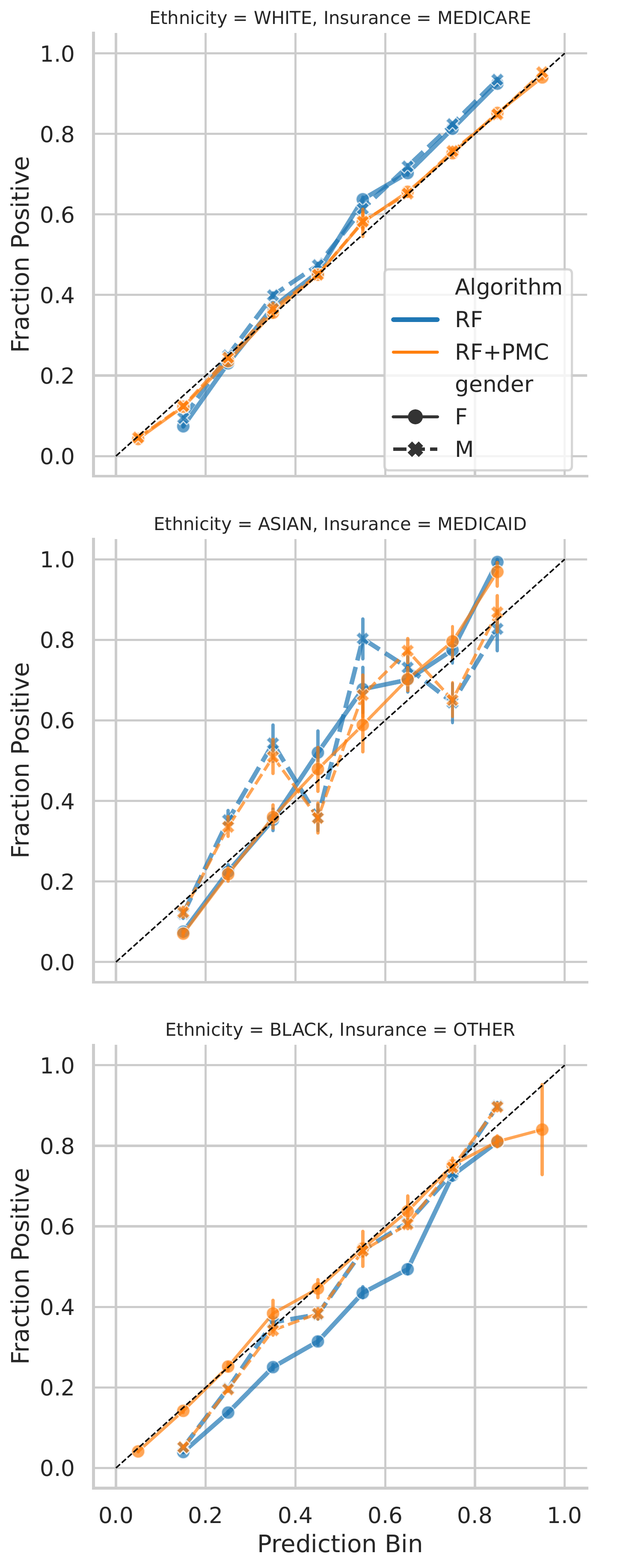}
    \caption{
        Calibration curves for RF models with and without PMCBoost postprocessing. 
    }
    \label{fig:cal}
\end{figure}

\begin{figure}
    \includegraphics[width=\columnwidth]{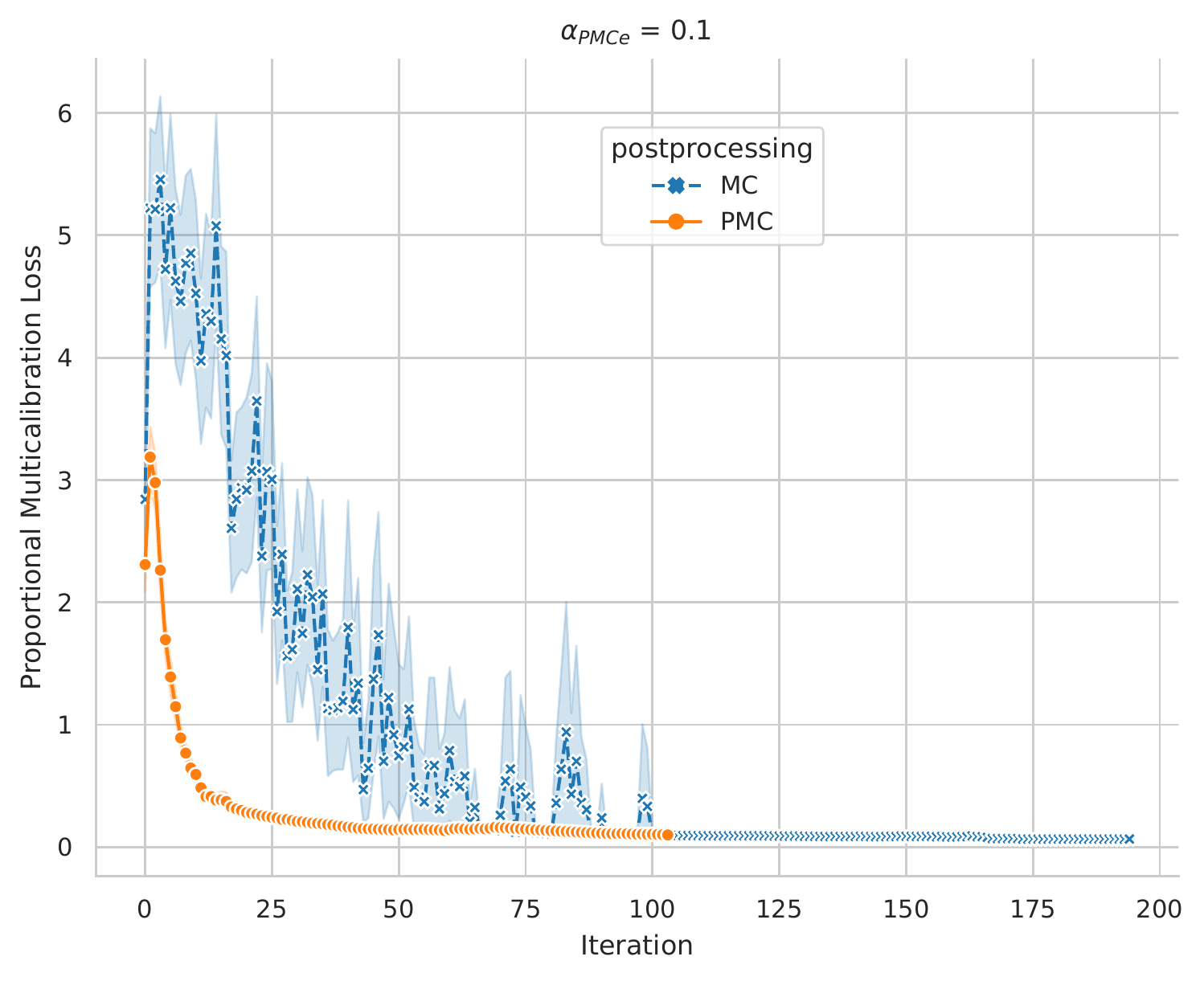}
    \caption{
        Number of iterations to convergence for PMCBoost and MCBoost when optimizing for the same value of $\alpha$-PMC ($\alpha_{PMCe}$). 
    }
    \label{fig:convergence}
\end{figure}

Results on the hospital admission prediction task are summarized in \cref{fig:mimic_results,fig:cal,fig:convergence,tbl:pct}. 
As \cref{tbl:pct} shows, PMCBoost has a small effect on predictive performance ($\Delta$AUROC $<$ 0.6\%) while improving DC loss and PMC loss by 10-81\% across LR, RF, and DNN models.    
Somewhat surprisingly, PMCBoost improves MC loss more effectively than MCBoost for both DNN (19\% versus 11\%) and RF models (18\% versus 6\%).
We observe that PMCBoost and MCBoost do not improve MC loss of LR models, although PMCBoost improves other metrics of fairness for LR (PMC loss and DC loss) without changing AUROC.  
In \cref{fig:cal}, we illustrate the calibration improvement of RF models using PMCBoost for three patient subgroups defined by the intersection of race/ethnicity, gender, and insurance type. 

\paragraph{Computational Cost}

\begin{table}
    \scriptsize
    \centering
    \caption{Median percent change in loss from the base model using MC- and PMCBoost.}
    \label{tbl:pct}
    \begin{tabular}{p{3.5em} p{1em} rrrr}
    \toprule
            &    &  AUROC & MC & PMC & DC \\
            & ML &  $\Delta$\% &  $\Delta$\% & $\Delta$\% & $\Delta$\%         \\
    \midrule
    MCBoost & DNN &  -0.58 &  -10.83 &   -16.27 &   -2.41 \\
            & LR &  -0.01 &   10.95 &    -9.50 &   -6.46 \\
            & RF &  -0.09 &   -5.91 &   -55.44 &  -13.88 \\
    PMCBoost & DNN &  -0.56 &  -19.49 &   -28.83 &   -9.91 \\
            & LR &   0.01 &   14.18 &   -37.90 &  -14.90 \\
            & RF &  -0.27 &  -17.53 &   -80.82 &  -23.52 \\
    \bottomrule
\end{tabular}
\end{table}

\begin{table}
        \centering
        \caption{
            For MCBoost and PMCBoost, we compare the average number of updates and wall clock time (s) taken to train for the equivalent values of $\alpha$-PMC. 
        }
        \label{tbl:time}
        \scriptsize
\begin{tabular}{lllrr}
        \toprule
           &       &     &  Iterations &  Time (s) \\
        ML & $\alpha_{PMCe}$ & Postprocessing &            &            \\
        \midrule
        LR & 0.001 & MC &     2094.0 &      678.0 \\
           &       & PMC &      263.0 &      320.0 \\
           & 0.010 & MC &      631.0 &      223.0 \\
           &       & PMC &      153.0 &      202.0 \\
        RF & 0.001 & MC &     1996.0 &      556.0 \\
           &       & PMC &      323.0 &      234.0 \\
           & 0.010 & MC &      713.0 &      197.0 \\
           &       & PMC &      256.0 &      167.0 \\
        \bottomrule
        \end{tabular}
\end{table}

As mentioned in \cref{s:pmcboost}, MCBoost may be configured to satisfy $\alpha$-PMC by setting $\alpha$ to a much smaller value, although this may take an excessive number of steps. 
To better understand this trade-off, we empirically compared MC- and PMCBoost by the number of steps required for each to reach their best performance in \cref{fig:convergence}. 
On average, MCBoost requires approximately 5x more updates to achieve similar performance on PMC loss as PMCBoost, due to its dependence on very small values of $\alpha$. 
Wall clock times scale similarly, as detailed in \cref{tbl:time}.

\paragraph{Sensitivity Analysis} 
In \appendixref{s:app:results}, we look at detailed performance comparisons of MCBoost and PMCBoost over values of $\alpha$ and group definitions in \cref{fig:auroc,fig:mcloss,fig:pmcloss}. 
We observe that, while low values of $\alpha$ for MCBoost improve its PMC loss performance, PMCBoost typically performs as well or better, particularly for larger values of $\alpha$, and does so in fewer iterations.

\section{Discussion and Conclusion}

In this paper we have analyzed multicalibration through the lens of sufficiency and differential calibration to reveal the sensitivity of this metric to correlations between outcome rates and group membership. 
We have proposed a measure, PMC, that alleviates this sensitivity and attempts to capture the ``best of both worlds" of MC and sufficiency. 
PMC provides equivalent percentage calibration protections to groups regardless of their risk profiles, and in so doing, bounds a model's differential calibration. 
We provide an efficient algorithm for learning PMC predictors by postprocessing a given risk prediction model.   
On a real-world and clinically relevant task (admission prediction), we have shown that post-processing three types of models with PMC leads to better performance across all three fairness metrics, with a small impact on predictive performance. 

Our preliminary analysis suggests PMC can be a valuable metric for training fair algorithms in resource allocation contexts. 
Future work could extend this analysis on both the theoretical and practical side. 
On the theoretical side, the generalization properties of the PMC measure should be established and its sample complexity quantified, as \citet{roseMachineLearningPrediction2018} did with MC. 
Additional extensions of PMC could establish a bound on the accuracy of PMC-postprocessed models in a similar vein to work by \citet{kimMultiaccuracyBlackboxPostprocessing2019} and \citet{hebert-johnsonMulticalibrationCalibrationComputationallyIdentifiable2018}. 
On the empirical side, future works should benchmark PMC on a larger set of real-world problems, and explore use cases in more depth. 
In addition, user studies could be employed to validate whether proportionally multicalibrated models do in fact instill more trust in decision-makers compared to baseline models. 

\acks{
    E.L. was partially supported by the Eunice Kennedy Shriver National Institute of Child Health and Human Development (NICHD) grant 5 T32 HD40128-19. 
    W.G.L. was partially supported by the National Institutes of Health (NIH) National Library of Medicine grant R00-LM012926. 
}

\newpage

\bibliography{bills_refs,references,manual}

\appendix

\section{Appendix}
\label{s:app}
In this section, we include additional comparisons to related work, additional definitions, proofs to the theorems in the main text, and additional experimental details. 
The code to reproduce the figures and experiments is available here: \repo.  

\subsection{Related Work}
\label{s:app:related}
\paragraph{Definitions of Fairness}
There are myriad ways to measure fairness that are covered in more detail in other works~\citep{barocasFairnessMachineLearning2019,chouldechovaFrontiersFairnessMachine2018,castelnovoZooFairnessMetrics2021}.
We briefly review three notions here. 
The first, \textit{demographic parity}, requires the model's predictions to be independent of patient demographics ($A$). 
Although a model satisfying demographic parity can be desirable when the outcome should be unrelated to sensitive attributes~\citep{fouldsAreParityBasedNotions2020}, it can be unfair if important risk factors for the outcome are associated with those attributes~\citep{hardtEqualityOpportunitySupervised2016a}.
For example, it may be more fair to admit socially marginalized patients to a hospital at a higher rate if they are assessed less able to manage their care at home. 
Furthermore, if the underlying rates of illness vary demographically, requiring demographic parity can result in a healthier patients from one group being admitted more often than patients who urgently need care. 

When the base rates of admission are expected to differ demographically, we can instead ask that the model's errors be balanced across groups. 
One such notion is \textit{equalized odds}, which states that for a given $Y$, the model's predictions should be independent of $A$. 
Satisfying equalized odds is equivalent to having equal FPR and FNR for every group in $A$. 

When the model is used for patient risk stratification, as in the target use case in this paper, it is important to consider a model's calibration for each demographic group in the data. 
Because risk prediction models influence who is prioritized for care, an unfairly calibrated model can systematically under-predict risk for certain demographic groups and result in under-allocation of patient care to those groups. 
Thus, guaranteeing group-wise calibration via an approach such as multicalibration also guarantees fair patient prioritization for health care provision. 
In some contexts, risk predictions are not directly interpreted, but only used to \textit{rank} patients, which in some contexts is sufficient for resource allocation.
Authors have proposed various ways of measuring the fairness of model rankings, for example by comparing AUROC between groups~\citep{kallusAssessingAlgorithmicFairness2020}.  

\paragraph{Approaches to Fairness}
Many approaches to achieving fairness guarantees according to demographic parity, equalized odds and its relaxations have been proposed~\citep{ dworkFairnessAwareness2012,hardtEqualityOpportunitySupervised2016a, berkConvexFrameworkFair2017,jiangIdentifyingCorrectingLabel2019a,kearnsPreventingFairnessGerrymandering2018}. 
When choosing an approach, is important to carefully weigh the relative impact of false positives, false negatives, and miscalibration on patient outcomes, which differ by use case.
When group base rates differ (i.e., group-specific positivity rates), \emph{equalized odds and calibration by group cannot both be satisfied}~\citep{kleinbergInherentTradeoffsFair2016}. 
Instead, one can often equalized multicalibration while satisfying relaxations of equalized odds such as \emph{equalized accuracy}, where $Accuracy = \mu TPR+(1-\mu)(1-FPR)$ for a group with base rate $\mu$. 
However, to do so may require denigrating the performance of the model on specific groups~\citep{chouldechovaFairPredictionDisparate2017a,pleissFairnessCalibration2017}, which is unethical in our context. 

As mentioned in the introduction, we are also motivated to utilize approaches to fairness that 1) dovetail well with intersectionality theory, and 2) provide privacy guarantees. 
Most work in the computer science/ machine learning space does not engage with the broader literature on socio-cultural concepts like intersectionality, which we see as a gap that makes adoption in real-world settings difficult~\citep{hannaCriticalRaceMethodology2020}. 
One exception to this statement is differential fairness~\citep{fouldsIntersectionalDefinitionFairness2019}, a measure designed with intersectionality in mind. 
In addition to being a definition of fairness that provides equal protection to groups defined by intersections of protected attributes, models satisfying $\epsilon$-differential fairness also satisfy $\epsilon$-pufferfish privacy. 
This privacy guarantee is very desirable in risk prediction contexts, because it limits the extent to which the model reveals sensitive information to a decision maker that has the potential to influence their interpretation of the model's recommendation. 
However, prior work on differential fairness has been limited to using it to control for demographic parity, which is not an appropriate fairness measure for our use case~\citep{fouldsAreParityBasedNotions2020}.

Multicalibration has inspired several extensions, including relaxations such as multiaccuracy~\citep{kimMultiaccuracyBlackboxPostprocessing2019}, low-degree multicalibration~\citep{gopalanLowDegreeMulticalibration2022}, and extensions to conformal prediction and online learning~\citep{jungMomentMulticalibrationUncertainty2021,guptaOnlineMultivalidLearning2021}. 
Noting that multicalibration is a guarantee over mean predictions on a collection of groups $\C$, \cite{jungMomentMulticalibrationUncertainty2021} propose to extend multicalibration to higher-order moments (e.g., variances), which allows one to estimate a confidence interval for the calibration error for each category. 
\cite{guptaOnlineMultivalidLearning2021} extend this idea and generalize it to the online learning context, in which an adversary chooses a sequence of examples for which one wishes to quantify the uncertainty of different statistics of the predictions.  
Recent work has also utilized higher order moments to ``interpolate" between the guarantees provided by multiaccuracy, which  only requires accuracy in expectation for groups in $\C$, and multicalibration, which requires accuracy in expectation at each prediction interval~\citep{kimMultiaccuracyBlackboxPostprocessing2019}. 
Like proportional multicalibration (\definitionref{def:PMC}), definitions of multicalibration for higher order moments provide additional criteria for quantifying model performance over many groups; in general, however, much of the focus in other work is on statistics for uncertainty estimation. 
Like these works, one may view our proposal for proportional multicalibration as alternative definition of what it means to be multicalibrated. 
The key difference is that proportional multicalibration measures the degree to which multicalibration depends on differences in outcome prevalence between groups, and in doing so provides guarantees of pufferfish privacy and differential calibration.  

\cite{dworkLearningOutcomesEvidenceBased2019} study the relation of fair rankings to multicalibration, and, in a similar vein to differential fairness measures, formulate a fairness measure for group rankings using the relations between pairs of groups.  
However, these definitions are specific to the ranking relation between the groups, whereas differential calibration cares only about the outcome differential (conditioned on model predictions) between pairs of groups. 

\subsubsection{Differential Fairness}
\label{s:app:DF}

DF was explicitly defined to be consistent with the social theoretical framework of \emph{intersectionality}. 
This framework dates back as early as the social movements of the '60s and '70s \citep{collins_intersectionality_2020} and  was brought into the academic mainstream by pioneering work from legal scholar Kimberlé Crenshaw~\citep{crenshawDemarginalizingIntersectionRace1989,crenshaw_mapping_1991} and sociologist Patricia Hill Collins~\citep{collins_black_1990}. 
Central to intersectionality is that hierarchies of power and oppression are structural elements that are fundamental to our society. 
Through an intersectional lens, these power structures are viewed as interacting and co-constituted, inextricably related to one another.
To capture this viewpoint, DF~\citep{fouldsIntersectionalDefinitionFairness2019} constrains the differential of a general data mechanism among all pairs of groups, where groups are explicitly defined as the intersections of protected attributes in $\A$.

\begin{definition}[$\epsilon$-differential fairness]
    \citep{fouldsIntersectionalDefinitionFairness2019}
    \label{def:DF}
    Let $\Theta$ denote a set of distributions and let $x \sim \theta$ for $\theta \in \Theta$.
    A mechanism $M(x)$ is $\eps$-differentially fair with respect to ($\C$,$\Theta$) 
    for all $\theta \in \Theta$ with $x \sim \theta$, and $m \in Range(M)$ if, 
    for all $(S_i,S_j) \in \C \times \C$ where $P(S_i|\theta)>0$, $P(S_j|\theta)>0$,
    \begin{equation}\label{eq:eDF} 
    e^{-\eps}\leq\frac{P_{M,\theta}(M(x)=m|S_i,\theta)}{P_{M,\theta}(M(x)=m|S_j,\theta)} \leq e^{\eps}
    \end{equation}
\end{definition}
\begin{definition}[Pufferfish Privacy]\label{def:puff}
    Let the collection of subsets $\C$ represent sets of secrets.  
    A mechanism $M({x})$ is $\epsilon$-\emph{pufferfish private} \citep{kiferPufferfishFrameworkMathematical2014} with respect to $(\C, \Theta)$ if for all $\theta \in \Theta$ with ${x} \sim \theta$, for all secret pairs $(S_i,S_j) \in \C \times \C$ and $y \in \mbox{Range}(M)$,
    \begin{equation}
    e^{-\epsilon} \leq \frac{P_{M, \theta}(M(x) = y| S_i, \theta)}{P_{M, \theta}(M(x) = y|S_j, \theta)}\leq e^\epsilon \mbox{ ,} \label{def:pufferfish}
    \end{equation}
    when $S_i$ and $S_j$ are such that  $P(S_i|\theta) > 0$, $P(S_j|\theta) > 0$.
\end{definition}
\paragraph{Note on pufferfish and differential privacy}
Although \cref{eq:eDF} is notable in its similarity to differential privacy~\citep{dwork2009differential}, they differ in important ways. 
Differential privacy aims to limit the amount of information learned about any one individual in a database by computations performed on the data (e.g. $M(x)$). 
Pufferfish privacy only limits information learned about the group membership of individuals as defined by $\C$. 
\cite{kiferPufferfishFrameworkMathematical2014} describe in detail the conditions under which these privacy frameworks are equivalent. 

\paragraph{Efficiency Property}
\label{s:df-efficient}
\cite{fouldsIntersectionalDefinitionFairness2019} also define an interesting property of $\eps$-differential fairness that allows guarantees of higher order (i.e., marginal) groups to be met for free; the property is given in \appendixref{s:app:def}. 

\begin{definition}[Efficiency Property] 
    \citep{fouldsIntersectionalDefinitionFairness2019} 
    \label{def:inter}
    Let $M(x)$ be an $\eps$-differentially fair mechanism with respect to $(\mathcal{C},\Theta)$. 
    Let the collection of subsets $\mathcal{C}$ group individuals according to the Cartesian product of attributes $A \subseteq \A$.  
    Let $\G$ be any collection of subsets that groups individuals by the Cartesian product of attributes in $A'$, where $A' \subset A$ and $A' \neq \emptyset$.  
    Then $M(x)$ is $\eps$-differentially fair in $(\G,\Theta)$.
\end{definition}

The authors call this the ``intersectionality property", although in practice it guarantees the reverse: if a model satisfies $\epsilon$-DF for the low level (i.e. intersectional) groups in $\C$, then it satisfies $\epsilon$-DF for every higher-level (i.e. marginal) group. 
For example, if a model is ($\epsilon$)-differentially fair for intersectional groupings of individuals by race and sex, then it is $\epsilon$-DF for the higher-level race and sex groupings as well.
Whereas the number of intersections grows exponentially as additional attributes are protected~\citep{kearnsPreventingFairnessGerrymandering2018}, the number of total possible subgroupings grows at a larger combinatorial rate: for $p$ protected attributes, we have $\sum_{k=1}^p{ \binom{p}{k} m_a^k}$ groups, where $m_a$ is the number of levels of attribute $a$. 
\paragraph{Limitations}
To date, analysis of DF for predictive modeling has been limited to defining $R(x)$ as the mechanism, which is akin to asking for \emph{demographic parity}.
Under demographic parity, one requires that model predictions be independent from group membership entirely, and this limits the utility of it as a fairness notion.
Although a model satisfying demographic parity can be desirable when the outcome should be unrelated to $\C$~\citep{fouldsAreParityBasedNotions2020}, it can be unfair if important risk factors for the outcome are associated with demographics~\citep{hardtEqualityOpportunitySupervised2016a}. 
For example, if the underlying rates of an illness vary demographically, requiring demographic parity can result in a healthier patients from one group being admitted more often than patients who urgently need care. 

\subsection{Proofs for Theorems in the Main Text}
\label{s:proof}

\paragraph{\theoremref{thm:MCtoDC}}
\label{proof:MCtoDC}
\textit{
    \Paste{thm:MCtoDC}
}
\begin{proof}
    Let $r = \ERr$ and $p^* = \Epstarr$.
    $\alpha$-MC guarantees that $r - \alpha \leq p^* \leq r + \alpha$ for all groups $S \in \C$ and predictions $r \in [0,1]$. 
    Plugging these lower and upper bounds into \cref{eq:DC}, we observe that the lower bound on $\eps$-DC for $R(x)$ is given by 
    $
        {\frac{r+\alpha}{r-\alpha} }  \leq e^{\eps} 
    $.  
    The maximum of the left-hand side for a fixed $\alpha$ occurs at the smallest value of $r$; therefore $R(x)$ satisfies
    $
        \ln \frac{r_{min} + \alpha}{r_{min} - \alpha} \leq \eps. 
    $
    By switching the numeratator and denominator we obtain the minimum differential and the left-hand side constraint from \definitionref{def:DC}, i.e.  
    $
        e^{-\eps} \leq \frac{r_{min} - \alpha}{r_{min} + \alpha} . 
    $
    Thus $R(x)$ is $\left( \ln \frac{r_{min}+\alpha}{r_{min}-\alpha} \right)$-differentially calibrated.
\end{proof}

\paragraph{\theoremref{thm:PMCtoDC}}
\label{proof:PMCtoDC}
\textit{
    \Paste{thm:PMCtoDC}
}
\begin{proof}
    Let $r = \ERr$ and $p^* = \Epstarr$.
    If $R(x)$ satisfies $\alpha$-PMC (\definitionref{def:PMC}), then 
    $ r/(1 + \alpha) \leq p^* \leq r/(1 - \alpha)$. 
    Solving for the upper bound on $\eps$-DC, we immediately have
    $\eps  \geq \ln \frac{r(1+\alpha)}{r(1-\alpha)} \geq \ln \frac{1+\alpha}{1-\alpha} $.
\end{proof}

\paragraph{\theoremref{thm:PMCtoMC}}
\label{proof:PMCtoMC}
\textit{
    \Paste{thm:PMCtoMC}
}
\begin{proof}
    To distinguish the parameters, let $R(x)$ be a model satisfying $\delta$-PMC. 
    Let $r = \ERr$ and $p^* = \Epstarr$.
    Then  $ r/(1 + \delta) \leq p^* \leq r/(1 - \delta) $. 
    We solve for the upper bound on $\alpha$-MC from \definitionref{def:MC} for the case when $p^* > r$. 
    This yields
    \begin{align*}
    \alpha &\leq p^* - r \\
           &\leq \frac{r}{1-\delta} - r \\
           &= r\frac{\delta}{1-\delta} \\
           &\leq \frac{\delta}{1-\delta}
           .
    \end{align*}
    We can also solve for the lower bound on $\alpha$-MC from \definitionref{def:MC} for the case when $p^* < r$. 
    This yields
    \begin{align*}
    \alpha &\leq r - p^* \\
           &\leq r - \frac{r}{1+\delta} \\
           &= r \frac{\delta}{1+\delta} \\
           &\leq \frac{\delta}{1+\delta}
           .
    \end{align*}
    For any $\delta > 0$, $\frac{\delta}{1-\delta} > \frac{\delta}{1+\delta}$. 
    Therefore the first case ($p^* > r$) limits the multicalibration of $R(x)$. 
\end{proof}

\paragraph{Proposition \ref{prop:alg}}\label{s:proof:alg}
\textit{
    \Paste{prop:alg}
}
\begin{proof}

    We show that \algorithmref{alg:PMC} converges using a potential function argument~\citep{bansalPotentialfunctionProofsGradient2019}, similar to the proof techniques for the MC boosting algorithms in \cite{hebert-johnsonMulticalibrationCalibrationComputationallyIdentifiable2018,kimMultiaccuracyBlackboxPostprocessing2019}. 
    Let $p^*_i$ be the underlying risk, $R_i$ be our initial model, and $R'_i$ be our updated prediction model for individual $i \in S_r$, where $S_r = \{x | x \in S, R(x) \in I\}$ and $(S,I) \in \C \times \Lambda_{\lambda}$.   
    We use $p^*$, $R$, and $R'$ without subscripts to denote these values over $S_r$. 
    We cannot easily construct a potential argument using progress towards ($\alpha$,$\lambda$)-PMC, since its derivative is undefined at $\EpstarI$=0.  
    Instead, we analyze progress towards the difference in the $\ell_2$ norm at each step. 

    \begin{align}
        & ||p^*-R|| - ||p^*-R'||  
        \nonumber\\
        & = \sum_{i \in S_r}{ (p_i^* - R_i)^2 } - \sum_{i \in S_r}{ (p_i^* - \text{squash}(R_i+\Delta r))^2 } 
        \nonumber \\
        & \geq   \sum_{i \in S_r}{\left( (p_i^* - R)^2 - (p_i^* - (R_i+\Delta r))^2 \right) } 
        \nonumber\\
        & = \sum_{i \in S_r}{\left( 2p_i^* \Delta r - 2R_i \Delta r - \Delta r^2 \right) } 
        \nonumber\\
        &= 2 \Delta r \sum_{i \in S_r}{\left( p_i^* - R_i \right)} - |S_r|\Delta r^2 \label{eq:del}
    \end{align}

    From \algorithmref{alg:PMC} we have 
    $$
    \Delta r =  \frac{1}{|S_r|}\sum_{i \in S_r}{( p_i^* - R_i )}
    $$
    Substituting into~\cref{eq:del} gives
    \begin{align*}
        ||p^*-R|| - ||p^*-R'|| &\geq |S_r|{\Delta r}^2 \\
    \end{align*}
    We know that $|S_r| \geq \alpha \lambda \gamma N$, and that the smallest update $\Delta r$ is $\alpha \rho$. 
    Thus, 
    \begin{align*}
        ||p^*-R|| - ||p^*-R'|| &\geq \alpha^3 \rho^2 \lambda \gamma N \\
    \end{align*}
    Since our initial loss, $|| p^* - R||$, is at most $N$, \algorithmref{alg:PMC} converges in at most $O(\frac{1}{\alpha^3 \rho^2 \lambda \gamma})$ updates for category $S_r$. 

    To understand the total number of steps, including those without updates, we consider the worst case, in which only a single category $S_r$ is updated in a cycle of the for loop (if no updates are made, the algorithm exits). 
    Since each repeat consists of at most $|C|/\lambda$ loop iterations, this results in $O(\frac{|C|}{\alpha^3 \lambda^2 \rho^2 \gamma})$ total steps. 
\end{proof}
\subsection{Extended Theoretical Analysis}
\label{s:app:theory}

\paragraph{Illustrating Relationships between Definitions}
\label{s:app:illustrate}

\begin{figure*}
    \centering
    \includegraphics[width=0.75\textwidth]{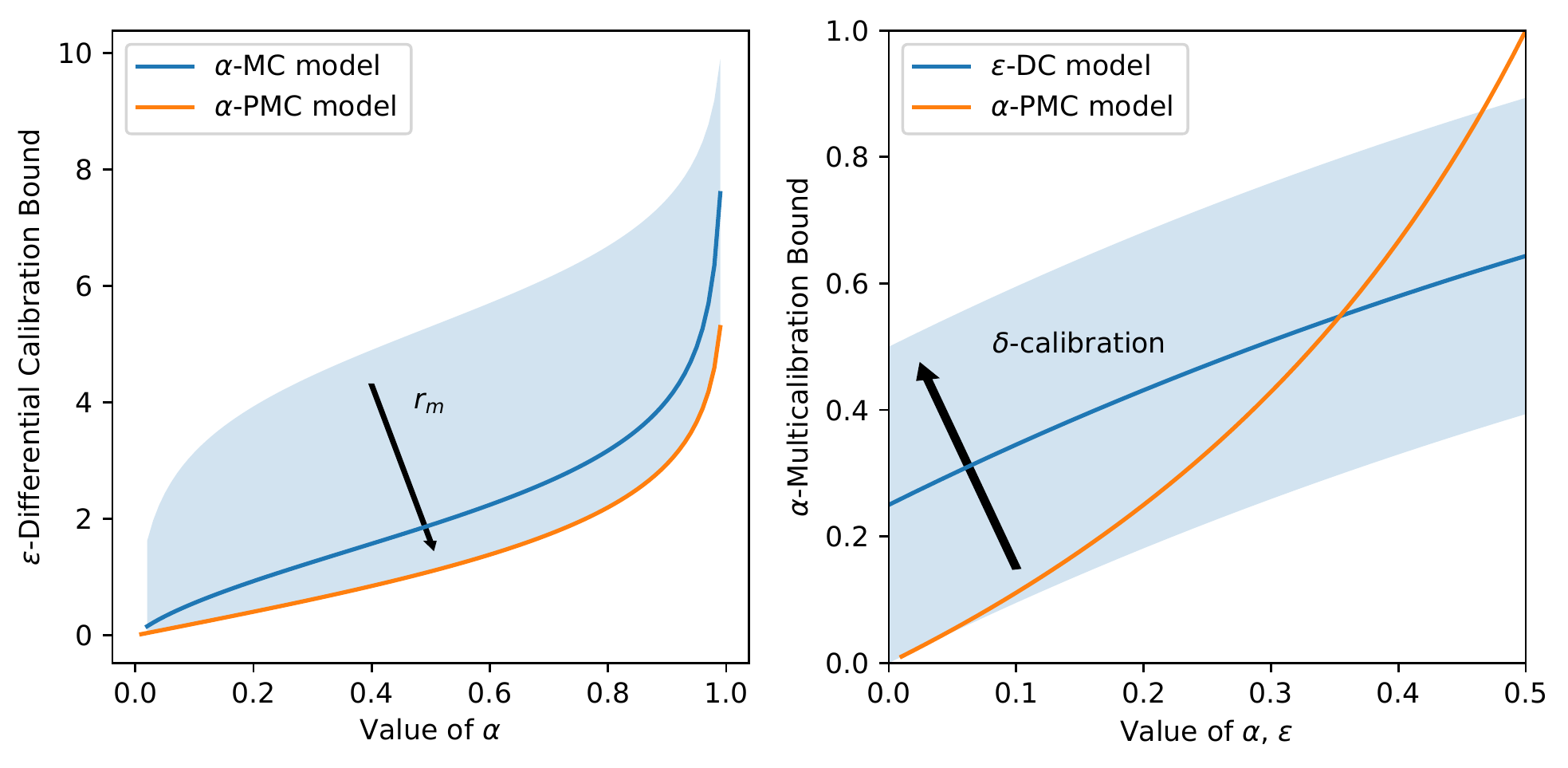}
    \caption{
        A comparison of $\eps$-DC, $\alpha$-MC, and $\alpha$-PMC in terms of their parameters $\alpha$ and $\epsilon$. 
    In both panes, the x value is a given value of one metric for a model, and the y axis is the implied value of the other metric, according to \theoremref{thm:DCtoMC}-\theoremref{thm:PMCtoMC}. 
    The left filled area denotes the dependence of the privacy/DC of $\alpha$-multicalibrated models on the minimum risk interval, $r_{min} \in [0.01, 1.0]$. 
    The right filled area denotes the dependence of the MC of $\epsilon$-differentially calibrated models on their overall calibration, $\delta \in [0.0, 0.5]$.
    $\alpha$-PMC does not have these sensitivities. 
    }
    \label{fig:params}
\end{figure*}

\cref{fig:params} shows how the definitions of MC, DC, and PMC are related.  
In each subplot, the x and y coordinates map the guarantee from one metric (x axis) to the implied guarantee in the other metric (y axis). 

The right panel of \cref{fig:params} illustrates this relation in comparison to the DC-MC relationship described in \appendixref{s:app:thm}, \theoremref{thm:DCtoMC}. 
At small values of $\epsilon$ and $\alpha$ and when the model is perfectly calibrated overall, $\alpha$-PMC and $\epsilon$-DC behave similarly. 
However, given $\delta>0$, $\epsilon$-differentially calibrated models suffer from higher MC error than proportionally calibrated models when $\alpha$-PMC $< 0.3$. 
The right graph also illustrates the feasible range of $\alpha$ for $\alpha$-PMC is $0 < \alpha < 0.5$, past which it does not provide meaningful $\alpha$-MC. 
The steeper relation between $\alpha$-PMC and MC may have advantages or disadvantages, depending on context. 
It suggests that, by optimizing for $\alpha$-PMC, small improvements to this measure can result in relatively large improvements to MC; conversely, $\epsilon$-DC models that are well calibrated may satisfy a lower value of $\alpha$-MC over a larger range of $\epsilon$. 

\subsubsection{Discretization}
\label{s:app:discrete}
To clarify and simplify our analysis, we work mainly with the continuous versions of multicalibration and proportional multicalibration, under the assumption that minimizing the discretized versions (i.e., binning $R(x)$) will translate to low values of the continous version. 
In this section we provide detailed bounds on the continuous versions of PMC and DC that are implied by the discretized versions. 

First, we will formally define two different discretization schemes. 
The first, $\lambda$-discretization, defines equally spaced bins on the interval $[0,1]$, as follows.

For ensuring multiplicative closeness under PMC, it can be useful to instead discretize the prediction bins so that the bins are equally spaced on a log scale. 
We define such a discretization below. 

\begin{definition}[$(\lambda,\rho)$-geometric discretization.]
    \label{def:log_discretization}
    Let $\lambda \in [0,1], \rho \in [0,1]$. 
    The \emph{$(\lambda,\rho)$-geometric discretization} of $[0,1]$ is denoted by a set of intervals, 
$\Lambda_{\lambda}^{\rho} = \Set{ \Set{I_j}_{j=0}^{1/\lambda -1}}$, where  
$
I_j = [ \rho^{(1-j\lambda)}, \rho^{(1-j\lambda-\lambda)} ) .
$
\end{definition}

\citep{hebert-johnsonMulticalibrationCalibrationComputationallyIdentifiable2018} define a discretized version of MC in which $R(x)$ is binned according to a discretization parameter, $\lambda$: 
\begin{definition}[$(\alpha,\lambda)$-multicalibration]
\label{def:MCdiscrete}
Let $\C \subseteq \text{2}^{\X}$ be a collection of subsets of $\X$. 
For any $\alpha, \lambda > 0$, 
a predictor $R$ is \emph{$(\alpha,\lambda)$-multicalibrated} on $\C$
if,  
for all $I \in \Lambda_{\lambda}$ 
and $S \in \C$ where $P_D(R \in I |x \in S) \geq \alpha \lambda $,
$$ \card{ \EpstarI - \ERI} \le \alpha .$$
\end{definition}

\citep{hebert-johnsonCalibrationComputationallyIdentifiableMasses2018} establish that $(\alpha, \lambda)$-multicalibrated models are at most $(\alpha+\lambda)$-multicalibrated.
In an analagous fashion, we show below that $(\alpha,\lambda)$-PMC implies $(\alpha+\lambda/\rho)$-PMC for bins defined by a $\lambda$-discretization. 
When using a $(\lambda,\rho)$-geometric discretization, $(\alpha, \lambda)$-PMC implies $(\alpha \rho^{-\lambda} + \rho^{-\lambda} - 1)$-PMC, which can be a tighter bound than the former. 

\begin{claim}
    \label{claim:abs}
    Define $\rho, \alpha, \lambda>0$ and  
    let $\C \subseteq \text{2}^{\X}$ be a collection of subsets of $\X$. 
    Let $\EpstarI \geq \rho$ for all $S \in \C$ and $I \in \Lambda_{\lambda}$.  
    Let $R(x)$ be a model satisfying $(\alpha,\lambda)$-proportional multicalibration. 
    Then $R(x)$ is at most $(\alpha + \frac{\lambda}{\rho})$-proportionally multicalibrated.
\end{claim}
\begin{proof}

    By \definitionref{def:PMCdiscrete}, $R(x)$ satisfies
    $$
        \frac{  \card{ \EpstarI - \ERI }   }
             {  \EpstarI   }   
             \le  \alpha
    $$
    for categories $(S,I) \in \C \times \Lambda_{\lambda}$ satisfying $P_D(R(x) \in I | x \in S) \geq \alpha \lambda$. 
    Given $1/\lambda$ bins, the subset where $P_D(R(x) \in I | x \in S) < \alpha \lambda$ has a size of at most $\alpha |S|$. 
    Therefore there is a subset $|S'| \geq (1-\alpha)|S|$ where for all $r \in \Lambda_{\lambda}$, 
    $\alpha$-PMC (\definitionref{def:PMC}) is satisfied.

    Let $\delta$ be the constaint on $\delta$-PMC. 
    Let $p^* = \Epstarr$ and $r = \ERr$. 
    Consider the case $r > p^*$ and let $\alpha = (r - p^*)/p^*$. 
    $\lambda$-discretization shifts $r$ by at most $\lambda$. 
    Let
    $$
    \delta \leq (r + \lambda - p^*)/p^*
    $$
    Substituting $r \leq \alpha p^* + p^*$ yields
    $$
    \delta \leq \alpha + \frac{\lambda}{p^*}
    $$
    Plugging in $\rho$ as the minimum of $p^*$, we complete the proof.

\end{proof}

The term $\frac{\lambda}{\rho}$ can be potentially large when $\rho < \lambda$. 
One way to avoid this issue is to make the change in $R(x)$ between bins scale with $R(x)$ using \definitionref{def:log_discretization}.  
What makes \definitionref{def:log_discretization} different from $\lambda$-discretization is that the intervals are a multiplicative, rather than additive, distance apart. 
Hence, for a given $r \in [0,1]$, a model satisfying $(\alpha,\lambda)$-PMC can have its prediction shift by at most a factor of $\rho^{-\lambda}$. 
This leads us to the following claim. 

\begin{claim}
    \label{claim:log}
    Define $\rho, \alpha, \lambda>0$ and  
    let $\C \subseteq \text{2}^{\X}$ be a collection of subsets of $\X$. 
    Let $\EpstarI \geq \rho$ for all $S \in \C$ and $I \in \Lambda_{\lambda}$.  
    Let $R(x)$ be a model satisfying $(\alpha,\lambda)$-proportional multicalibration. 
    Given a $(\lambda,\rho)$-geometric discretization, $R(x)$ is at most $(\alpha \rho^{-\lambda} + \rho^{-\lambda} - 1)$-proportionally multicalibrated.
\end{claim}
\begin{proof}

    By \definitionref{def:PMCdiscrete}, $R(x)$ satisfies
    $$
        \frac{  \card{ \EpstarI - \ERI }   }
             {  \EpstarI   }   
             \le  \alpha
    $$
    for categories $(S,I) \in \C \times \Lambda_{\lambda}$ satisfying $P_D(R(x) \in I | x \in S) \geq \alpha \lambda$. 
    Given $1/\lambda$ bins, the subset where $P_D(R(x) \in I | x \in S) < \alpha \lambda$ has a size of at most $\alpha |S|$. 
    Therefore there is a subset $|S'| \geq (1-\alpha)|S|$ where for all $r \in \Lambda_{\lambda}$, 
    $\alpha$-PMC (\definitionref{def:PMC}) is satisfied.

    Let $\delta$ be the constaint on $\delta$-PMC. 
    Let $p^* = \Epstarr$ and $r = \ERr$. 
    Consider the case $r > p^*$ and let $\alpha = (r - p^*)/p^*$, i.e. the tight bound. 
    $(\lambda,\rho)$-geometric discretization shifts $r$ by at most a factor of $\rho^{-\lambda}$. 
    This implies
    $$
    \delta \leq (r \rho^{-\lambda} - p^*)/p^*
    $$
    Substituting $r = \alpha p^* + p^*$ yields
    $$
    \delta \leq \alpha \rho^{-\lambda}+ \rho^{-\lambda} - 1.
    $$

\end{proof}

\begin{figure}
    \centering
    \includegraphics[width=\columnwidth]{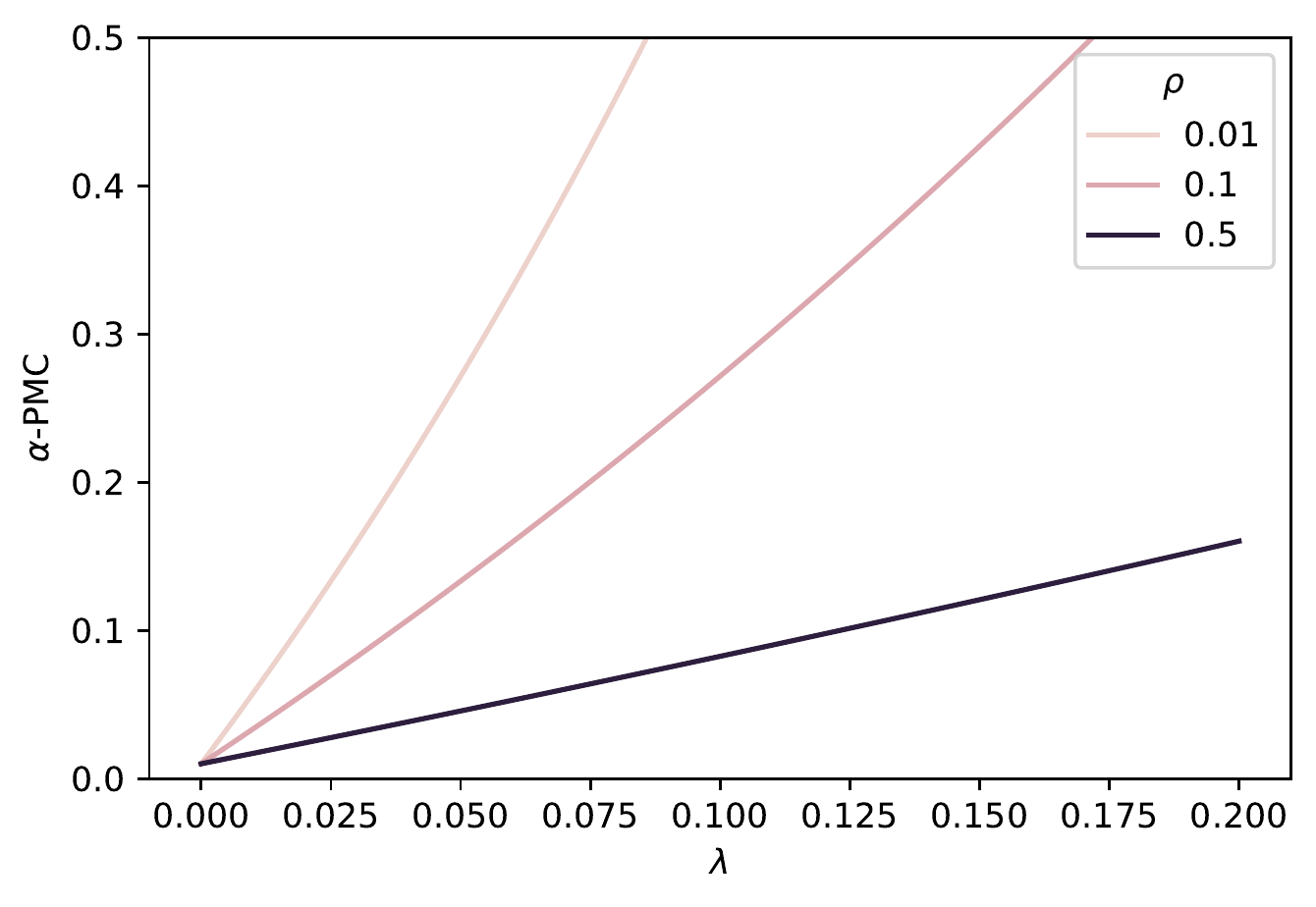}
    \caption{
        Relationship between $(\alpha,\lambda)$-PMC and $\alpha$-PMC given a geometric discretization. 
        Illustrated for $(\alpha,\lambda)$-PMC=0.1, for various values of $\rho$ and $\lambda$.
    }
    \label{fig:discrete}
\end{figure}

We illustrate the relationship between $(\alpha,\lambda)$-PMC and $\alpha$-PMC given a geometric discretization in \cref{fig:discrete}, which quantifies the relationship for different settings of $\lambda$ and $\rho$. 

\subsubsection{Additional Definitions}
\label{s:app:def}

\begin{definition}[$\lambda$-discretization.]
    \label{def:discretization}
    Given $\lambda$ $\in [0,1]$, 
the \emph{$\lambda$-discretization} of $[0,1]$ is denoted by a set of intervals, 
$
\Lambda_{\lambda} = \Set{ \{I_j\}_{j=0}^{1/\lambda -1}}
$, where  
$I_j = [ j\lambda, (j+1)\lambda ) ] $.
\end{definition}



\begin{definition}[$(\alpha,\lambda)$-PMC]
    \label{def:PMCdiscrete}
    A model $R(x)$ is $(\alpha,\lambda)$-\emph{proportionally multicalibrated} with respect to a collection of subsets $\C$ if, 
    for all $S \in \C$ and $I \in \Lambda_\lambda$ satisfying $P_D(R \in I | x \in S) \geq \alpha \lambda$,
    \begin{equation}\label{eq:PMCdiscrete}
        \frac{  \card{ \EpstarI - \ERI }   }
             {  \EpstarI   }   
             \le  \alpha.
    \end{equation}
\end{definition}

The following loss functions are empirical analogs of the definitions of $MC$, $PMC$, and $DC$, and are used in the experiment section to measure performance. 

\begin{definition}[MC loss]
    \label{def:mcloss}
    Let $\mathcal{D} = \Set{(y,x)_i}_{i=0}^{N} \sim D$, and let $\alpha, \lambda, \gamma > 0$.
    Define a collection of subsets $\C \in 2^{\X}$ such that for all $S \in \C, |S| \geq \gamma N$.
    Let $S_I = \Set{x: R(x) \in I, x \in S}$ for $(S,I) \in \C \times \Lambda_\lambda$.
    Define the collection $\mathcal{S}$ containing all $S_I$ satisfying $S_I \geq \alpha \lambda N$.
    The MC loss of a model $R(x)$ on $\mathcal{D}$ is 
\[
    \max_{S_I \in \mathcal{S}}{
    \frac{1}{|S_I|}
    \card{\sum_{i \in S_I}{ y_i } - \sum_{i \in S_I}{ R_i }}
}
\]
\end{definition}

\begin{definition}[PMC loss]
    \label{def:pmcloss}
    Let $\mathcal{D}=\Set{(y,x)_i}_{i=0}^{N} \sim D$, and let $\alpha, \lambda, \gamma, \rho > 0$.
    Define a collection of subsets $\C \in 2^{\X}$ such that for all $S \in \C, |S| \geq \gamma N$.
    Let $S_I = \Set{x: R(x) \in I, x \in S}$ for $(S,I) \in \C \times \Lambda_\lambda$.
    Define the collection $\mathcal{S}$ containing all $S_I$ satisfying $S_I \geq \alpha \lambda N$.
    Let $\frac{1}{|S_I|}\sum_{i \in S_I}{ y_i } \geq \rho$.
    The PMC loss of a model $R(x)$ on $\mathcal{D}$ is 
\[
    \max_{S_I \in \mathcal{S}}{
    \frac{
        \card{\sum_{i \in S_I}{ y_i } - \sum_{i \in S_I}{ R_i }}
    }
    {
        \sum_{i \in S_I}{ y_i }
    }
}
\]
\end{definition}

\begin{definition}[DC loss]
    \label{def:dcloss}
    Let $\mathcal{D} = \Set{(y,x)_i}_{i=0}^{N} \sim D$, and let $\alpha, \lambda, \gamma > 0$.
    Define a collection of subsets $\C \in 2^{\X}$ such that for all $S \in \C, |S| \geq \gamma N$.
    Given a risk model $R(x)$ and prediction intervals $I$, 
    Let $S_I = \Set{x: R(x) \in I, x \in S}$ for $(S,I) \in \C \times \Lambda_\lambda$.
    Define the collection $\mathcal{S}$ containing all $S_I$ satisfying $S_I \geq \alpha \lambda N$.
    The DC loss of a model $R(x)$ on $\mathcal{D}$ is 
\[
    \max_{(S_I^a,S_I^b) \in \mathcal{S} \times \mathcal{S}}{
    \log{\card{
        \frac{1}{|S_I^a|} \sum_{i \in S_I^a}{ y_i } - \frac{1}{|S_I^b|}\sum_{j \in S_I^b}{ y_j }
    }}
}
\]
\end{definition}

\subsection{Additional Theorems}\label{s:app:thm} 

\subsubsection{Differentially calibrated models with global calibration are multicalibrated}
Here we show that, under the assumption that a model is globally calibrated (satisfies $\delta$-calibration), models satisfying $\eps$-DC are also multicalibrated. 


\begin{theorem}\label{thm:DCtoMC}
    Let R(x) be a model satisfying ($\eps$,$\lambda$)-DC and $\delta$-calibration. 
    Then $R(x)$ is ($1-e^{-\eps}+\delta$, $\lambda$)-multicalibrated. 
\end{theorem}
\begin{proof}

From~\cref{eq:DC} we observe that $\eps$ is bounded by the two groups with the largest and smallest group- and prediction- specific probabilities of the outcome. 
Let $I_M$ be the risk stratum maximizing $(\eps,\lambda)$-DC, and let $p_n = \max_{S \in \C} P_D(y|R \in I_M, x \in S)$ and $p_d = \min_{S \in \C} P_D(y|R \in I_M,x \in S)$. 
These groups determine the upper and lower bounds of $\eps$ as $e^{-\eps} \leq p_d/p_n$ and $p_n/p_d \leq e^{\eps}$. 

We note that $p_d \leq P_D(y|R \in I_M) \leq p_n$, since $P(y| R \in I_M) = \frac{1}{N} \sum_{S \in \C} |S| P_D(y|R \in I_M, x \in S)$, and $p_n$ and $p_d$ are the extreme values of $P(y|R\in I_M,x \in S)$ among $S$. 
So, $\alpha$-MC is bound by the group outcome that most deviates from the predicted value, which is either $p_n$ or $p_d$. 
Let $r = P_D( R|R \in I_M )$.
There are then two scenarios to consider:

\begin{enumerate}
    \item $ \alpha \leq | p_n - r | = p_n - r $ when $r \leq \frac{1}{2}(p_n + p_d)$; and
    \item $ \alpha \leq | p_d - r | = r - p_d $ when $r \geq \frac{1}{2}(p_n + p_d)$.
\end{enumerate} 

We will look at the first case. 
Let $p^*_r = P_D(y|R \in I_M)$. 
Due to $\delta$-calibration, $p^*_r - \delta \leq r \leq p^*_r + \delta$. 
Then 
\begin{align*}
    \alpha  &\leq p_n - r \\
            &\leq p_n - (p^*_r - \delta) \\
            &\leq p_n - p_d + \delta \\
            &= p_n (1-e^{-\eps}) + \delta\\
   \alpha   &\leq 1 - e^{-\eps} + \delta. 
\end{align*}

Above we have used the facts that $r \leq p^*_r - \delta$, $p^*_r \geq p_d$, $p_d \leq e^{-\eps}p_n$, and $p_n \leq 1$. 
The second scenario is complementary and produces the identical bound. 
\end{proof}
\theoremref{thm:DCtoMC} formally describes how $\delta$-calibration controls the baseline calibration error contribution to $\alpha$-MC, while $\eps$-DC limits the deviation around this value by constraining the (log) maximum and minimum risk within each category. 

\subsubsection{Multicalibrated models satisfy intersectional guarantees}

In contrast to DF, MC \citep{hebert-johnsonMulticalibrationCalibrationComputationallyIdentifiable2018} was not designed to explicitly incorporate the principles of intersectionality. 
However, we show that it provides an identical efficiency property to DF in the theorem below. 
Given an individual's attributes $x = (x_1,\;\dots,\;x_d)$, it will be useful to refer to subsets we wish to protect, e.g. demographic identifiers.
To do so, we define $\A = \Set{A_1,\;\dots,\;A_p}$, $p \leq d$, such that $A_1$ is the set of values taken by attribute $x_1$.  

\begin{theorem}\label{thm:intersectionalmc}
    Let the collection of subsets $\mathcal{C} \subseteq 2^\X$ define groups of individuals according to the Cartesian product of attributes $A \subseteq \A$.  
    Let $\G \in 2^\X$ be any collection of subsets that groups individuals by the Cartesian product of attributes in $A'$, where $A' \subset A$ and $A' \neq \emptyset$.  
    If $R(x)$ satisfies $\alpha$-MC on $\C$, then $R(x)$ is $\alpha$-multicalibrated on $\G$.
\end{theorem}

In proving \theoremref{thm:intersectionalmc}, we will make use of the following lemma. 

\begin{lemma}\label{lemma:express} 
    The $\alpha$-MC criteria can be rewritten as: for a collection of subsets $\C \subseteq \X$, $\alpha \in [0,1]$, and $r \in [0,1]$,
$$
\max_{c\in\C}\E_D [ y | R(x) = r, x \in c ]\leq r+\alpha 
$$
and
$$
\min_{c\in\C}\E_D [ y | R(x) = r, x \in c ] \ge r-\alpha
$$
\end{lemma}

\begin{proof}
The lemma follows from~\definitionref{def:MC}, and simply restates it as a constraint on the maximum and minimum expected risk among groups at each prediction level. 
\end{proof}

\begin{proof}[Proof of \theoremref{thm:intersectionalmc}]
%
    We use the same argument as \cite{fouldsIntersectionalDefinitionFairness2019} in proving this property for DF. 
Define $Q$ as the Cartesian product of the protected attributes included in $\A$, but not $\A'$. 
Then for any $(y,x) \sim D$,


\begin{align}
    & \max_{g\in \G} \E_D [ y | R(x) = r, x \in g] 
    \nonumber \\
    &= \max_{g \in \G}\sum_{q\in Q} \E_D [ y | R(x) = r, x \in g \cap q ]P[x \in q | x \in g]
    \label{eq:max1}
    \\
    &\leq \max_{g \in \G}\sum_{q\in Q} \max_{q'\in Q}\E_D [ y | R(x) = r, x \in g \cap q' ]P[x \in q | x \in g]
    \label{eq:max2}
    \\
    &= \max_{g \in \G}\max_{q'\in Q}\E_D [ y | R(x) = r, x \in g \cap q' ]
    \label{eq:max3}
    \\
    &= \max_{c\in \C} \E_D [ y | R(x) = r, x \in c ]
        .
    \label{eq:max4}
\end{align}

Moving from \cref{eq:max1} to \cref{eq:max2} follows from substituting the maximum value of $\E_D [ y | R(x) = r, x]$ for observations in the intersection of subsets in $\mathcal{G}$ and $Q$ which is the upper limit of the expression in \cref{eq:max1}. 
Moving from \cref{eq:max2} to \cref{eq:max3} follows from recognizing that the sum $P[x\in q|x \in g]$ for all subsets in $\mathcal{Q}$ is 1. 
Finally, moving from \cref{eq:max3} to \cref{eq:max4} follows from recognizing that the intersections of subsets in $\mathcal{G}$ and $\mathcal{Q}$ that satisfy \cref{eq:max3}, must define a subset of $\mathcal{C}$.
Applying the same argument, we can show that

$$
\min_{g\in \G}\E_D[y|R(x)=r,x\in g] \ge\min_{c\in \C} \E_D [ y | R(x) = r, x \in c ] 
.
$$
Substituting into \cref{lemma:express},
$$
\max_{g\in \G}\E_D [ y | R(x) = r, x \in g]\leq \alpha+r\\
$$
and
$$
\min_{g\in \G}\E_D [ y | R(x) = r, x \in g ] \ge r -\alpha
$$

or

$$
\card{\E_D [ y | R(x) = r, x \in g ]-r}\le{\alpha} 
$$
for all $g\in \G$. Therefore $R(x)$ is $\alpha$-multicalibrated with respect to $\G$.

\end{proof}

As a concrete example, imagine we have the protected attributes $A = \Set{ \text{race} \in \Set{B,W}, \text{gender} \in \Set{M,F}}$. 
According to \theoremref{thm:intersectionalmc}, $\C$ would contain four sets: \{$(B,M)$,$(B,F)$,$(W,M)$,$(W,F)$\}. 
In contrast, there are eight possible sets in $\G$: \{ $(B,M)$,$(B,F)$,$(W,M)$,$(W,F)$,$(B,*)$,$(W,*)$,$(*,M)$, $(*,F)$\}, where the wildcard indicates a match to either attribute. 
As noted in \appendixref{s:df-efficient}, the efficiency property is useful because the number of possible sets in $\G$ grows at a large combinatorial rate, rate as additional attributes are added; meanwhile $\C$ grows at a slower, yet exponential, rate. 
For an intuition for why this property holds, consider that the maximum calibration error of two subgroups is at least as large as the maximum expected error of those groups combined; e.g., the maximum calibration error in a higher order groups such as $(B,*)$ will be covered by the maximum calibration error in either $(B,M)$ or $(B,F)$.

\subsection{Additional Experiment Details}
\label{s:app:exp}
\paragraph{Models}
The deep neural network (DNN) was a five layer feed-forward NN with 100 units per layer and ReLU activations.  
We trained using an adam solver with a learning rate of 0.001, a batch size of 200 and used early stopping to terminate when a 10\% validation set did not improve for 10 epochs. 

Both LR and DNN models used median imputation and feature normalization as pre-processing steps. 
For the RF, we used the XGBoost implementation~\cite{chenXGBoostScalableTree2016} which handles missing data natively.  

\paragraph{Training}
Models were trained on a heterogenous computing cluster. 
Each training instance was limited to a single core and 4 GB of RAM. 
We conducted a full parameter sweep of the parameters specified in \cref{tbl:params}.
A single trial consisted of a method, a parameter setting from \cref{tbl:params}, and a random seed. 
Over 100 random seeds, the data was shuffled and split 75\%/25\% into train/test sets.
Results in the manuscript are summarized over these test sets. 

\paragraph{Code}
\label{s:code}

Code for the experiments is available here: \repo. 
Code is licensed under GNU Public License v3.0.
\paragraph{Data}

We make use of data from the \href{https://physionet.org/content/mimic-iv-ed/1.0/}{MIMIC-IV-ED} repository, version 1.0, to train admission risk prediction models~\citep{johnsonalistairMIMICIVED2021}.
This resource contains more than 440,000 ED admissions from Beth Isreal Deaconness Medical Center between 2011 and 2019. 
We preprocessed these data to construct an admission prediction task in which our model delivers a risk of admission estimate for each ED visitor after their first visit to triage, during which vitals are taken. 
Additional historical data for the patient was also included (e.g., number of previous visits and admissions). 
A list of features is given in \cref{tbl:features}.

\subsection{Additional Experimental Results}
\label{s:app:results}

\begin{figure*}
    \includegraphics[width=\textwidth]{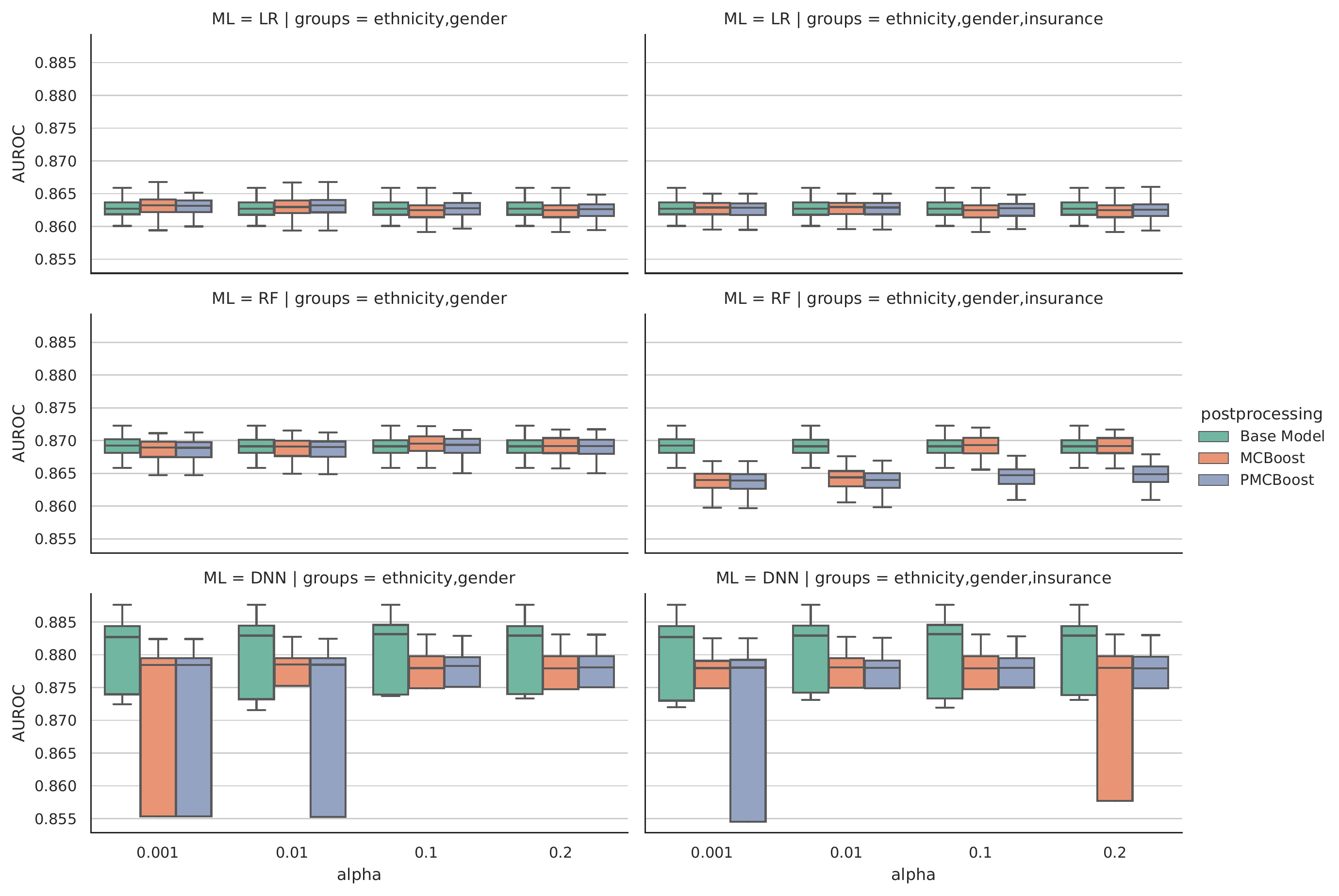}
    \caption{
        AUROC test performance versus $\alpha$ across experiment settings.
        Rows are different ML base models, and columns are different attributes used to define $\C$. 
        The color denotes the post-processing method. 
    }
    \label{fig:auroc}
\end{figure*}
\begin{figure*}
    \includegraphics[width=\textwidth]{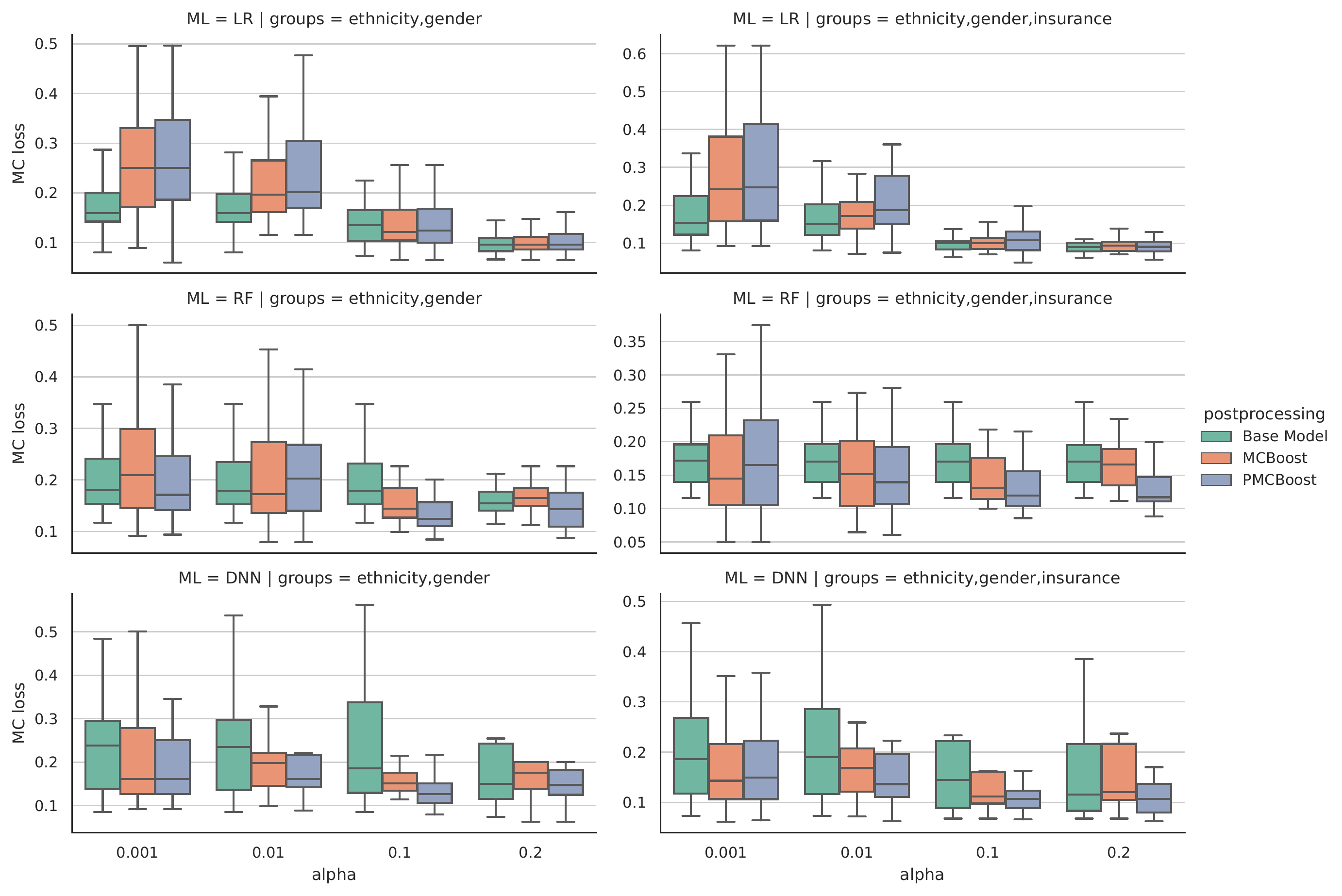}
    \caption{
        MC loss test performance versus $\alpha$ across experiment settings.
        Rows are different ML base models, and columns are different attributes used to define $\C$. 
        The color denotes the post-processing method. 
    }
    \label{fig:mcloss}
\end{figure*}
\begin{figure*}
    \includegraphics[width=\textwidth]{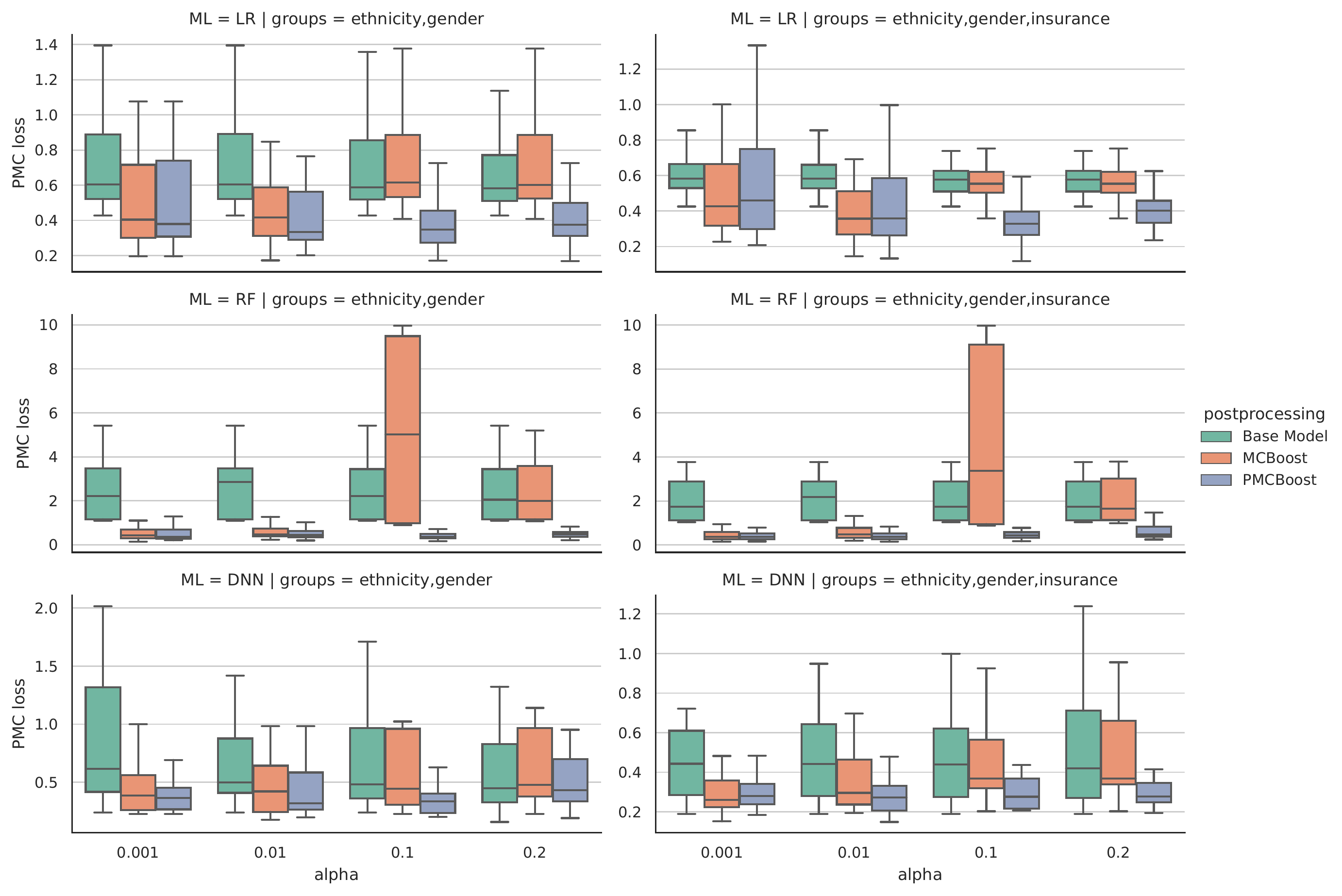}
    \caption{
        PMC loss test performance versus $\alpha$ across experiment settings.
        Rows are different ML base models, and columns are different attributes used to define $\C$. 
        The color denotes the post-processing method. 
    }
    \label{fig:pmcloss}
\end{figure*}
\begin{figure*}
    \includegraphics[width=\textwidth]{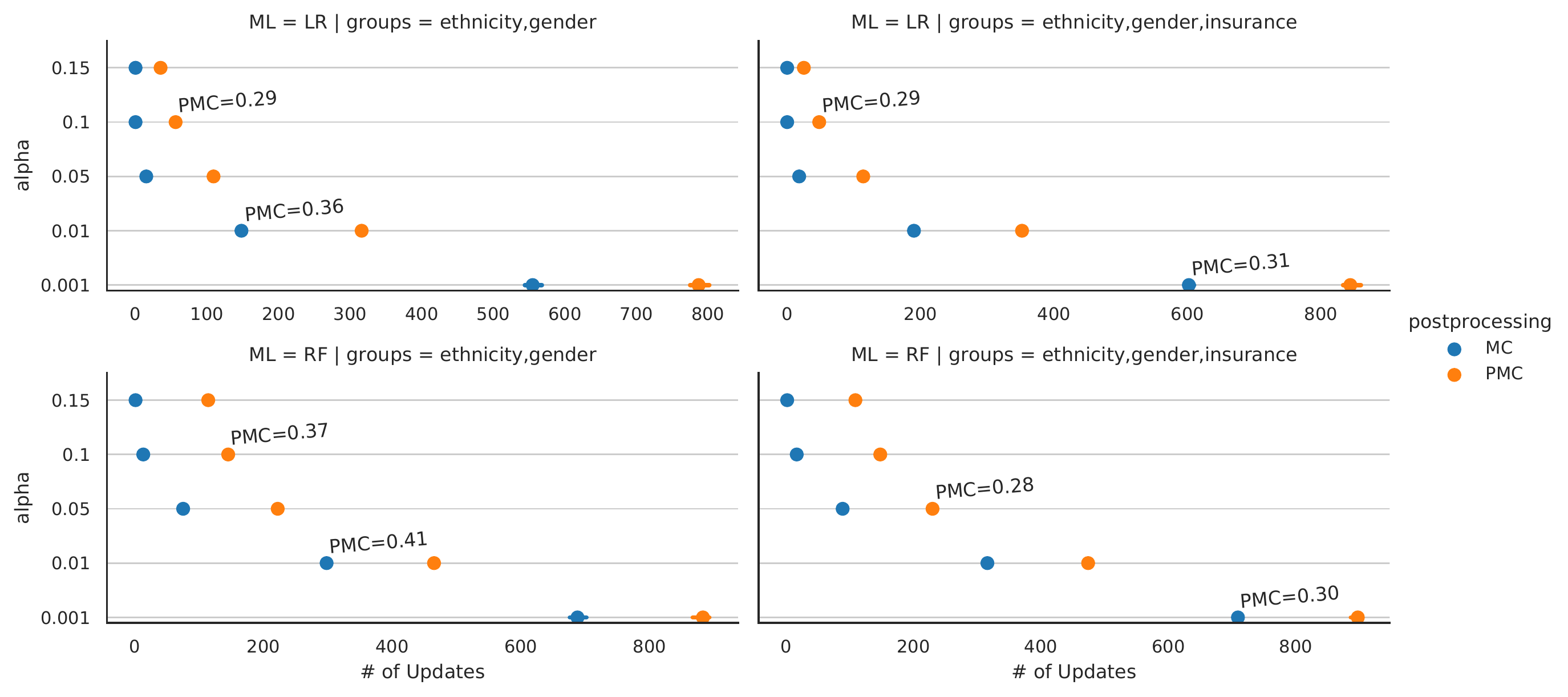}
    \caption{
        Number of post-processing updates by MC and PMC versus $\alpha$ across experiment settings.
        Rows are different ML base models, and columns are different attributes used to define $\C$. 
        The color denotes the post-processing method. 
        Each result is annotated with the median PMC loss for that method and parameter combination. 
    }
    \label{fig:updates}
\end{figure*}

\cref{tbl:params} lists a few parameters that may affect the performance of post-processing for both MC and PMC.  
Of particular interest when comparing MC versus PMC post-processing is the parameter $\alpha$, which controls how stringent the calibration error must be across categories to terminate, and the group definition ($A$), which selects which features of the data will be used to asses and optimize fairness. 
We look at the performance of MC and PMC postprocessing over values of $\alpha$ and group definitions in \cref{fig:auroc,fig:mcloss,fig:pmcloss}. 
Finally, we empirically compare MC- and PMC-postprocessing by the number of steps required for each to reach their best performance in \cref{fig:updates,tbl:time}. 

From \cref{fig:auroc}, it is clear that post-processing has a minimal effect on AUROC in all cases; note the differences disappear if we round to two decimal places. 
When post processing with RF, we do note a relationship between lower values of $\alpha$ and a very slight decrease in performance, particularly for MC-postprocessing. 

\cref{fig:mcloss,fig:pmcloss} show performance between methods on MC loss and PMC loss, respectively. 
In terms of MC loss, PMC-postprocessing tends to produce models with the lowest loss, at $\alpha$ values greater than 0.01. 
Lower values of $\alpha$ do not help MC-postprocessing in most cases, suggesting that these smaller updates may be overfitting to the post-processing data. 
In terms of PMC loss (\cref{fig:pmcloss}), we observe that performance by MC-postprocessing is highly sensitive to the value of $\alpha$.
For smaller values of $\alpha$, MC-postprocessing is able to achieve decent performance by these metrics, although in all cases, PMC-postprocessing generates a model with a better median loss value at some configuration of $\alpha$. 

We assess how many steps/updates MC and PMC take for different values of $\alpha$ in \cref{fig:updates}, and summarize empirical measures of running time in \cref{tbl:time}. 
On the figure, we annotate the point for which each post-processing algorithm achieves the lowest median value of PMC loss across trials. 
\cref{fig:updates} validates that PMC-postprocessing is more efficient than MC-postprocessing at producing models with low PMC loss, on average requiring 4.0x fewer updates to achieve its lowest loss on test. 
From \cref{tbl:time} we observe that PMC typically requires a larger number of updates to achieve its best performance on MC loss (about 2x wall clock time and number of updates), whereas MC-postprocessing requires a larger number of updates to achieves its best performance on PMC loss and DC loss, due to its dependence on very small values of $\alpha$. 
We accompany these results with the caveat that they are based on performance on one real-world task, and wall clock time measurements are influenced by the heterogenous cluster environment; future work could focus on a larger empirical comparison.

We quantify how often the use of each post-processing algorithm gives the best loss for each metric and trial in \cref{tbl:wins}. 
PMCBoost (\algorithmref{alg:PMC}) achieves the best fairness the highest percent of the time, according to DC loss (63\%), MC loss (70\%), and PMC loss (72\%), while MC-postprocessed models achieve the best AUROC in 88\% of cases.  
This provides strong evidence that, over a large range of $\alpha$ values, PMCBoost is beneficial compared to MCBoost.  

\begin{table}
    \centering
    \footnotesize
    \caption{
        The number of times each postprocessing method achieved the best score among all methods, out of 100 trials. 
    }
    \scriptsize
\begin{tabular}{lrrr}
\toprule
postprocessing &  Base Model &  MC &  PMC \\
metric   &             &     &      \\
\midrule
AUROC    &           5 &  88 &    6 \\
MC loss  &           8 &  21 &   70 \\
PMC loss &           0 &  27 &   72 \\
DC loss  &           0 &  36 &   63 \\
\bottomrule
\end{tabular}
    \label{tbl:wins}
\end{table}


\end{document}